\documentclass[sigconf]{acmart}
\AtBeginDocument{%
  }

\copyrightyear{2026}
\acmYear{2026}
\setcopyright{cc}
\setcctype{by}
\acmConference[MM '26]{Proceedings of the 34th ACM International Conference on Multimedia}{November 10--14, 2026}{Rio de Janeiro, Brazil}
\acmBooktitle{Proceedings of the 34th ACM International Conference on Multimedia (MM '26), November 10--14, 2026, Rio de Janeiro, Brazil}
\acmDOI{10.1145/3767308.3836205}
\acmISBN{979-8-4007-2213-4/2026/11}

\usepackage{multirow}
\usepackage{enumitem}
\begin{document}

\title{Every Packet Counts: Dispersing Information for Loss-Resilient Learned Image Compression}

\author{Yuhang Wei}
\authornotemark[1]
\email{1150501302@sjtu.edu.cn}
\orcid{0009-0001-2108-0703}
\affiliation{%
  \institution{Shanghai Jiao Tong University}
  \city{Shanghai}
  \country{China}}

\author{Chuqin Zhou}
\authornote{Both authors contributed equally to this research.}
\email{zhouchuqin@sjtu.edu.cn}
\affiliation{
  \institution{Shanghai Jiao Tong University}
  \city{Shanghai}
  \country{China}}

\author{Yibo Shi}
\email{shiyibo@huawei.com}
\affiliation{
  \department{Central Media Technology Institute}
  \institution{Huawei Technologies Ltd.}
  \city{Beijing}
  \country{China}}

\author{Jing Wang}
\email{wangjing215@huawei.com}
\affiliation{
  \department{Central Media Technology Institute}
  \institution{Huawei Technologies Ltd.}
  \city{Beijing}
  \country{China}}

\author{Guo Lu}
\authornote{Corresponding author.}
\email{luguo2014@sjtu.edu.cn}
\affiliation{
  \institution{Shanghai Jiao Tong University}
  \city{Shanghai}
  \country{China}}

\renewcommand{\shortauthors}{Yuhang Wei, Chuqin Zhou, Yibo Shi, Jing Wang, \& Guo Lu}

\begin{abstract}
Learned image compression (LIC) has achieved impressive rate-distortion performance. However, existing methods remain highly vulnerable to packet loss, a common challenge in satellite and emergency communications. This vulnerability stems from non-uniform information distribution at the packetization stage and sequential decoding dependencies at the entropy coding stage. We propose an end-to-end loss-resilient image compression scheme that addresses both. Before packetization, we introduce an Inter-Channel Redistribution (ICR) mechanism to redistribute channel energy, preventing critical information concentrating in a small subset of channels. Then, an Interleaved Channel Grouping (ICG) strategy partitions latent channels in a strided manner to disperse information across packets, with each packet kept within constrained sizes. To limit cascading errors from lost packets, we adopt a two-layer dual-branch autoregressive structure to shorten the dependency chain. Extensive experiments demonstrate that our method consistently outperforms existing approaches in both reconstruction quality and stability. At 20\% packet loss, it achieves an average PSNR gain of 1.84 dB over LossResilientLIC while reducing PSNR variance by an order of magnitude. Notably, trained under uniform random loss only, our model generalizes to bursty loss modeled by the Gilbert–Elliott channel, outperforming methods explicitly trained for such conditions.
\end{abstract}

\ccsdesc[500]{Computing methodologies~Image compression}
\keywords{Image Compression; Neural Network; Loss-Resilient}


\maketitle

\begin{figure}[h]
  \centering
  \includegraphics[width=\linewidth]{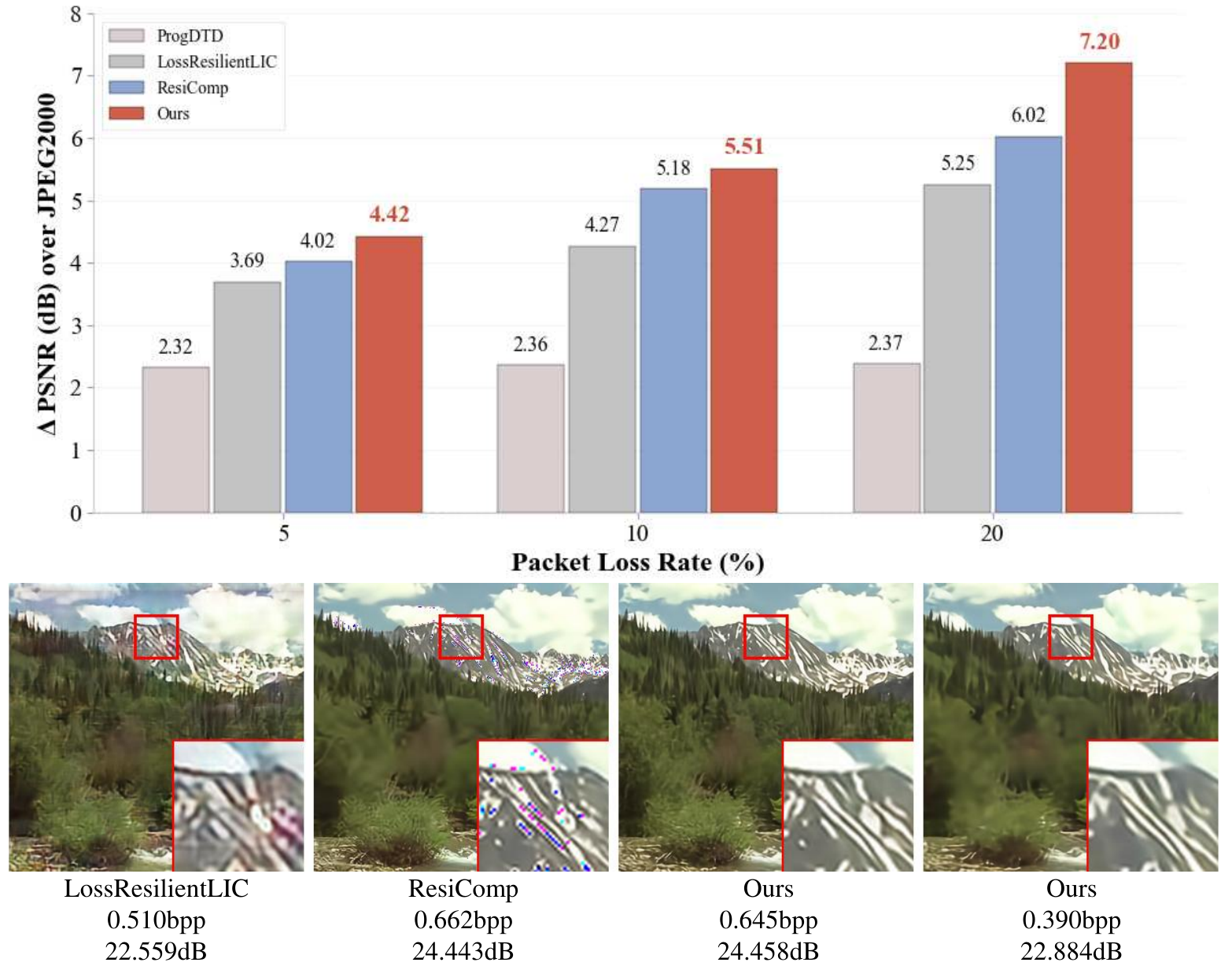}
  \caption{PSNR gains over JPEG2000 at medium bitrate under varying packet loss rates. Visual comparisons show that LossResilientLIC produces gray artifacts and ResiComp exhibits corrupted texture, whereas our method achieves higher PSNR at a lower bitrate with visually clean reconstruction.}
  \label{fig:first image with visualization}
  \Description{fig:first image}
\end{figure}

\section{Introduction}

The explosive growth of multimedia data has driven an increasing demand for efficient image and video transmission, spurring sustained research on image and video compression. Learned Image Compression (LIC) has achieved remarkable success in rate-distortion performance. Recent approaches~\cite{Cui_2021_CVPR_AG-VAE, Pan_22_ECCV_content, Duan_23_WACV_QRes-VAE, Feng_23_CVPR_NVTC, Li_23_NeurIPS_idempotent, Duan_23_TPAMI_QARV, Wu_25_AAAI_CLC, Zhou_25_AAAI_glic, Zhou_25_AAAI_controllable} have demonstrated substantial superiority over established standards such as JPEG~\cite{Wallace_1991_CACM_JPEG}, BPG~\cite{Bellard_2018_BPG}, and VVC~\cite{Bross_2021_TCSVT_VVC}. However, most existing LIC methods are developed and evaluated under the assumption of lossless transmission. In real-world deployment, particularly in satellite communications, emergency networks, and long-distance links, packet loss inevitably occurs due to network congestion, signal fluctuation, and timeouts. Under such scenarios, LIC methods suffer dramatic performance degradation, yet this critical challenge remains largely underexplored.

Deploying LIC under lossy channels faces challenges at two stages of the compression pipeline. At the \textbf{packetization stage}, existing methods adopt unequal packetization strategies that concentrate critical information in a small subset of packets. Loss of particular packets severely degrades reconstruction quality, even if all subsequent packets are received. At the \textbf{entropy coding stage}, a separate but equally critical issue arises. Entropy coding introduces sequential dependencies within each bitstream such that all preceding bits are required for arithmetic decoding, so a single lost segment can render the remainder of that bitstream undecodable. Autoregressive entropy models~\cite{He_2021_CVPR_Checkerboard, He_22_CVPR_ELIC, Jiang_23_MM_MLIC} further amplify this issue by introducing latent-wise dependencies. When earlier packets are lost, estimation errors propagate along the autoregressive chain, causing cascading quality degradation that worsens with longer dependency chains. Progressive coding approaches~\cite{Lee_22_CVPR_DPICT, Hojjat_23_CVPR_ProgDTD} share this vulnerability. Loss of early, high-priority packets disrupts all subsequent decoding. Both factors make reconstruction sensitive not just to how many packets are lost, but to \emph{which} ones are lost, causing unstable performance across different loss patterns.

Recent efforts have emerged to address packet loss within the LIC framework. LossResilientLIC~\cite{Sha_25_AAAI_LossResilientLIC} redistributes spatially adjacent latents across packets via spatial-channel rearrangement. However, its tail-drop training strategy concentrates critical information in leading transmitted packets, leaving the packetization-stage problem unresolved. To tackle the entropy-coding-stage dependency, it adopts a non-autoregressive architecture, which sacrifices entropy modeling capacity and overall compression performance, as depicted in Figure~\ref{fig:first image with visualization}. ResiComp~\cite{Wang_25_TCSVT_ResiComp}, built on masked visual token modeling, predicts lost tokens via a bidirectional transformer and performs well above 0.2~bpp using Intra Slice mode without entropy-coding-stage dependencies. Yet at low bitrates, it falls back to Layered mode with hierarchical autoregressive dependencies, where packet loss at any layer propagates downward and amplifies decoding errors. In both cases, fundamental issues persist that packet information remains unevenly distributed, leaving reconstruction vulnerable to the loss of specific packets.

To address challenges at both stages, we propose a loss-resilient image compression scheme that disperses information and minimizes quality impacts of packet loss. First, we introduce an Inter-Channel Redistribution (ICR) mechanism that leverages channel interaction to disperse channel energy before packetization, preventing energy concentration, while the Inverse Inter-Channel Redistribution (Inv-ICR) aggregates channel energy before decoding. Second, we design an Interleaved Channel Grouping (ICG) strategy that partitions latent channels in a strided manner so that each packet is of comparable importance, no significant degradation being caused by loss of any specific packet. Third, to limit cascading effects of entropy-coding-stage dependencies, we adopt a two-layer dual-branch autoregressive structure that shortens the dependency chain to two levels; with a training strategy that assigns low-criticality information to the second layer, this design bounds reconstruction degradation even when preceding packets are lost. Extensive evaluations on Kodak~\cite{Franzen_93_url_kodak} and CLIC~\cite{Toderici_20_clic} datasets demonstrate that our method achieves superior reconstruction quality and stability over both traditional and LIC-based approaches across varying packet loss rates and transmission environments. Our main contributions are summarized as follows:

\begin{itemize}[itemsep=1pt, topsep=-10pt, parsep=1pt, partopsep=12pt]
\item We design an Inter-Channel Redistribution mechanism to redistribute channel energy and an Interleaved Channel Grouping strategy to disperse packet information. They render the reconstruction insensitive to the loss of any specific packet.
\item We adopt a two-layer dual-branch autoregressive structure that eliminates first-layer intra-stream dependencies and limits cross-layer cascading effects, enabling robust entropy decoding under packet loss.
\item Experiments demonstrate that our method achieves state-of-the-art loss-resilient performance across multiple packet loss rates. At 20\% packet loss, our method achieves an average PSNR gain of 1.84~dB over LossResilientLIC with an order-of-magnitude reduction in variance. 
\end{itemize}

\begin{figure*}[t]
  \includegraphics[width=\textwidth]{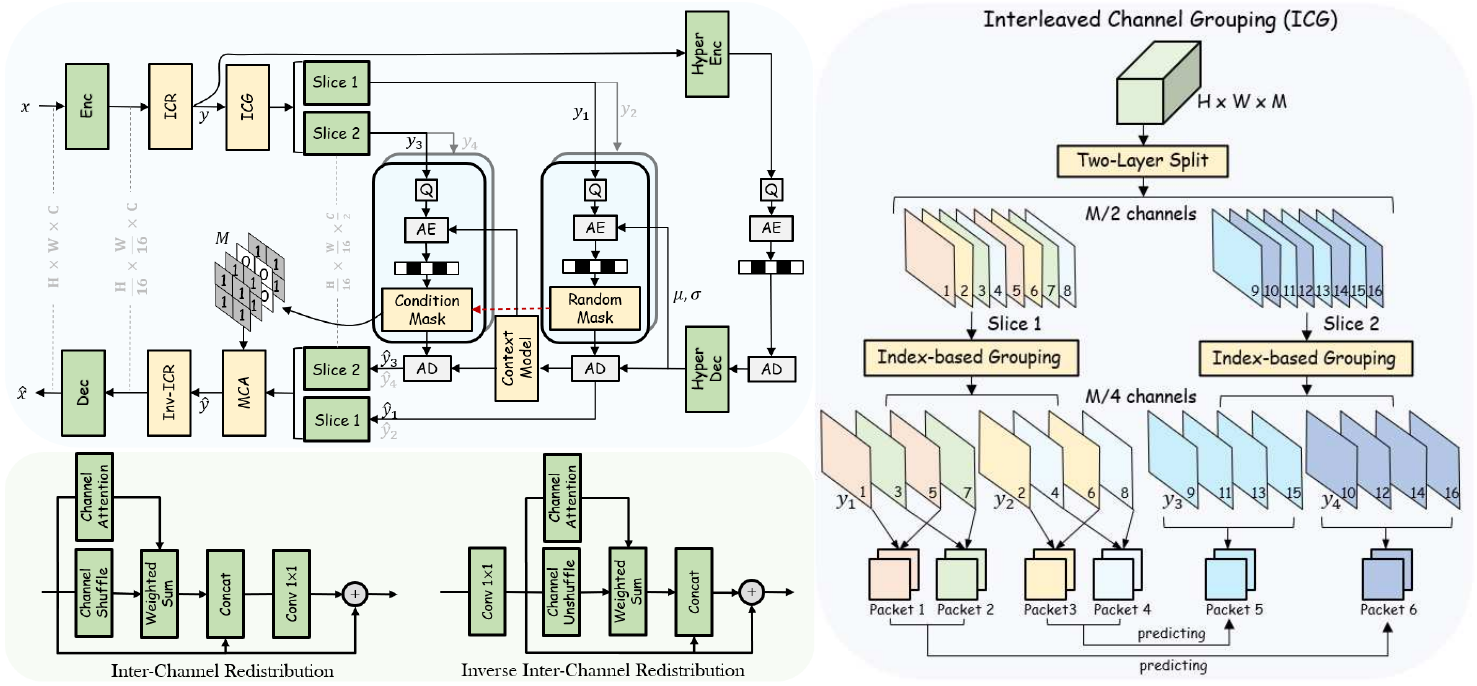}
  \caption{Overview of our proposed method. The Inter-Channel Redistribution (ICR) module shuffles and disperses channel information. Subsequently, the Interleaved Channel Grouping (ICG) module performs hierarchical splitting and packet encapsulation. Random mask simulates packet-level loss on packets during transmission, while Condition Mask generates masks for packets based on the random loss and autoregressive dependency. Before decoding, the Inverse Inter-Channel Redistribution (Inv-ICR) module unshuffles and aggregates channels.} 
  \Description{fig:network overview}
  \label{fig:network overview}
\end{figure*}

\section{Related Works}
\subsection{Learned Image Compression}
In recent years, Learned Image Compression (LIC) has achieved remarkable progress by leveraging deep neural networks. Theis \textit{et al.}~\cite{Theis_17_ICLR_lossy} and Ballé \textit{et al.}~\cite{Balle_2017_ICLR_E2E} pioneer end-to-end image compression frameworks based on variational autoencoders (VAEs), in which images are transformed into compact latent representations through learned analysis and synthesis transforms. To capture the scale priors of spatially adjacent elements, they~\cite{Balle_2018_ICLR_hyperprior} further introduce a hierarchical prior model, which has since become a foundational component in the field. To enhance the accuracy of entropy estimation, Minnen \textit{et al.}~\cite{Minnen_2018_NeurIPS_Joint} propose an autoregressive context module that conditions the distribution estimation of each latent element on its already-decoded neighbors, enabling more precise probabilistic modeling. Subsequent research has explored more principled loss functions~\cite{Ali_23_NeurIPS_Correlation, Li_25_ICLR_AuxT}, more sophisticated transform architectures~\cite{Cheng_2020_CVPR_DGML, Qin_24_arxiv_MambaVC, Feng_25_CVPR_LALIC, Chen_25_ICCV_KD, Zeng_25_CVPR_MambaIC, Chen_26_CVPR_GLIC, Chen_26_ICLR_CMIC} to eliminate redundancy, and more expressive context modules~\cite{Minnen_20_ICIP_CC, Jiang_23_ICMLW_mlicpp, Li_25_ICCV_HPCM} for accurate entropy modeling.

In addition, progressive coding approaches based on LIC have been explored to mitigate the long latency associated with high-delay transmission. DPICT~\cite{Lee_22_CVPR_DPICT} proposes a progressive compression algorithm supporting fine granular scalability, ensuring that the earlier packet encapsulates the most critical information. ProgDTD~\cite{Hojjat_23_CVPR_ProgDTD} introduces a progressive compression strategy that introduces no additional model parameters, employing a double-tail-drop mechanism to concentrate important information within leading filters.

\subsection{Traditional Loss-Resilient Coding}
During transmission, packet loss frequently occurs due to network congestion and timeout-induced delays. Retransmission, widely adopted in TCP, is a conventional countermeasure. However, it considerably increases user latency in long-distance communication due to acknowledgment delays and high round-trip times (RTTs). 

Forward Error Correction (FEC)~\cite{Gruber_17_CISS_deep, Rudow_23_NSDI_Tambur} is a widely used encoder-side approach to mitigating packet loss by introducing redundancy. Specifically, for $N$ source packets, additional $R$ redundant packets are generated to enable error verification and recovery at the receiver, accommodating loss rates up to $p = \frac{R}{N+R}$. Nevertheless, FEC entails a fundamental trade-off between bandwidth utilization and loss resilience. Under favorable network conditions, the redundant packets result in bandwidth inefficiency, whereas under adverse conditions where loss rates exceed $p = \frac{R}{N+R}$, decoding may fail. 

Post-processing Error Concealment (PEC) offers an alternative decoder-side strategy that requires no redundant injection. Techniques such as intra-mode macroblock encoding~\cite{Chu_98_TCSVT_detection} and flexible macroblock ordering~\cite{Ismaeil_00_ICIP_efficient} enable each transmitted packet to be decoded independently, ensuring that the loss or corruption of one packet does not affect decoding of others. However, this independence comes at costs of compression efficiency, as the encoder is precluded from exploiting redundancy across adjacent macroblocks.

\subsection{LIC-Based Loss-Resilient Coding}

Recent years have witnessed growing interest in loss-resilient compression built upon LIC. GRACE~\cite{Cheng_24_NSDI_GRACE} introduces an end-to-end video compression system that disperses information across multiple packets, incorporating optimistic encoding with dynamic state resynchronization for recovery. ProgDTD~\cite{Hojjat_23_CVPR_ProgDTD} achieves progressive coding via dual-drop training, enabling partial reconstruction under loss, but fails to accommodate loss of early, high-priority packets. LossResilientLIC~\cite{Sha_25_AAAI_LossResilientLIC} distributes spatially adjacent latent elements across packets via spatial-channel rearrangement, and restores missing elements through a Mask Conditional Aggregation (MCA) module. However, its training mechanism tends to concentrate critical information within leading packets of transmission sequences, inducing reconstruction quality highly sensitive to loss of specific packets. ResiComp~\cite{Wang_25_TCSVT_ResiComp} predicts lost tokens via masked visual token modeling, and systematically investigates loss-resilient performance across various contextual coding configurations. However, Layered mode employed at low bitrates exhibits poor loss resilience due to its hierarchical autoregressive dependencies, which amplify the impact of packet loss on subsequent decoding steps.

\begin{figure*}[t]
  \centering
  \begin{minipage}{0.48\linewidth}
    \centering
    \includegraphics[width=\linewidth]{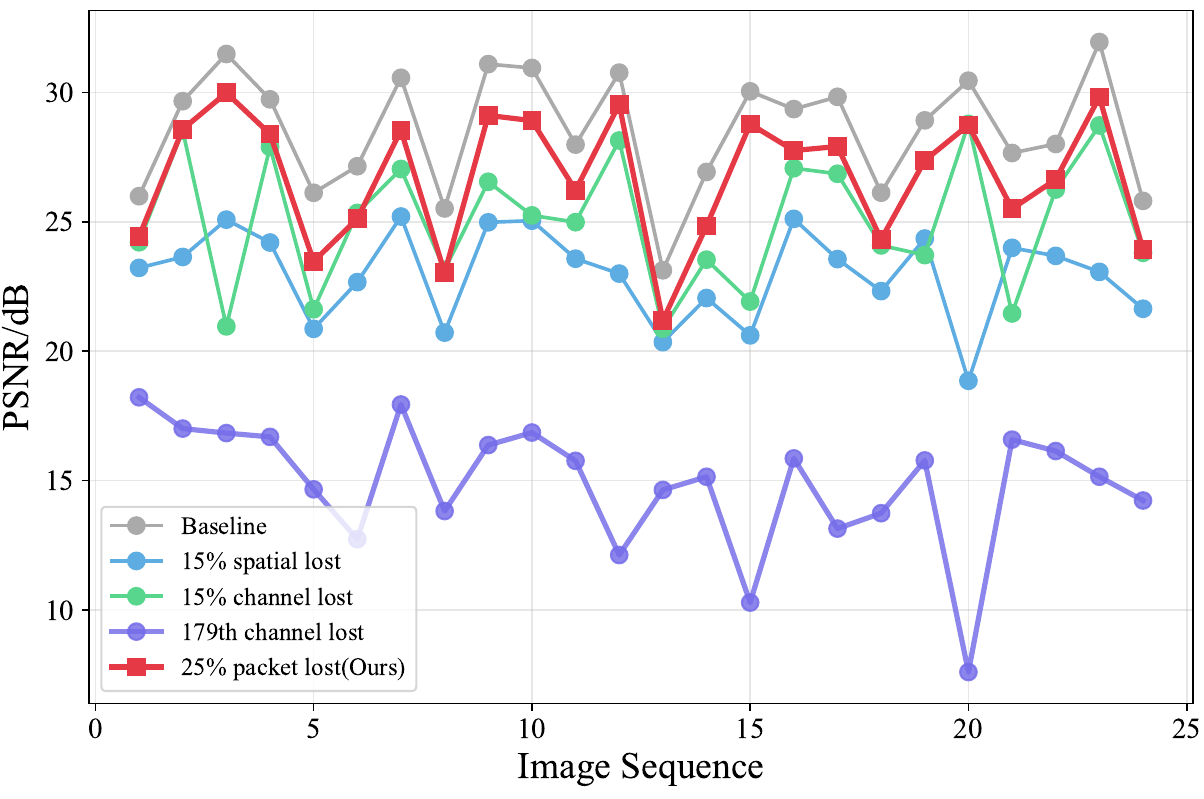}
    \caption*{(a) PSNR of reconstructed images with element loss.}
  \end{minipage}
  \hfill
  \begin{minipage}{0.48\linewidth}
    \centering
    \includegraphics[width=\linewidth]{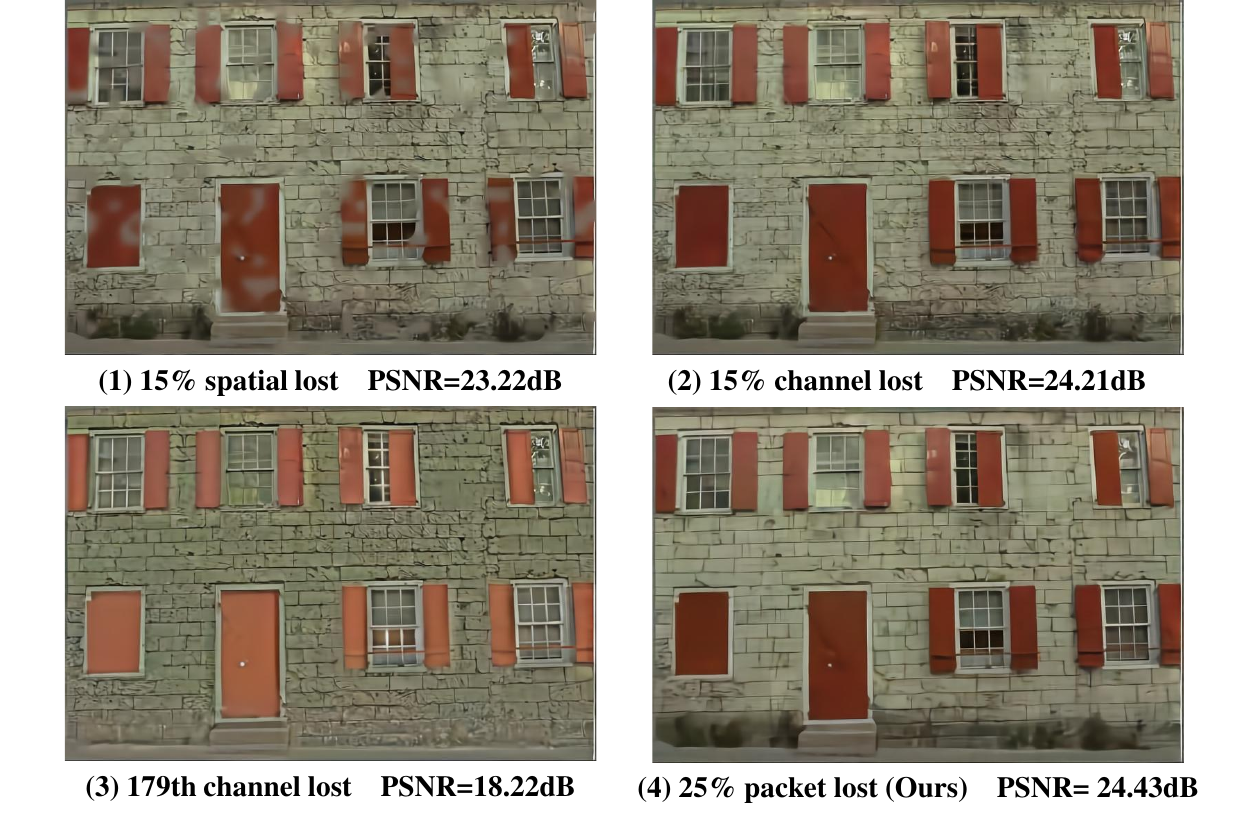}
    \caption*{(b) Visualization of reconstruction under different conditions.}
  \end{minipage}
  \caption{Analysis of information loss across spatial and channel dimensions. ``Baseline'' denotes the case without element loss, while ``Spatial'' and ``Channel'' refer to random element loss along the corresponding dimensions. The 179th channel corresponds to the channel with the highest energy. ``Ours'' represents the proposed model trained for loss resilience.}
  \Description{}
  \label{fig:spatial and channel}
\end{figure*}

\section{Method}
\subsection{Overview}
Our proposed framework is illustrated in Figure~\ref{fig:network overview}. The key design components introduced to mitigate packet loss are described below. The Inter-Channel Redistribution(ICR) mechanism leverages channel-based attention and channel rearrangement to effectively redistribute inter-channel energy. To prevent substantial quality fluctuations resulting from loss of any particular packet, we design an Interleaved Channel Grouping (ICG) strategy, dispersing critical information across different packets. Additionally, to shorten the dependency chain, we introduce a two-layer dual-branch autoregressive model, encapsulating each first-layer group as an independently decodable packet. It is worth noting that we assume $\hat{z}$ of the hyperprior branch to be lossless during transmission, as loss of the hyperprior stream would corrupt distribution estimates for decoding $\hat{y}$, rendering correct reconstruction infeasible. To adapt the model to packet loss during inference, the training incorporates a structured masking strategy with both random and conditional masking. Specifically, when packets in the first-layer are lost, masking is propagated to second-layer packets with autoregressive dependencies on affected packets. The overall model is optimized with respect to a rate-distortion objective, formulated as follows:
\begin{equation}
\mathcal{L} = \mathcal{R} + \lambda \cdot \mathcal{D} = \mathcal{R}\left(\hat{y}\right) + \lambda_z \cdot \mathcal{R}\left(\hat{z}\right) + \lambda \cdot \mathcal{D}\left(x, \hat{x}\right)
\label{eq:loss}
\end{equation}
where $\mathcal{R}(\hat{y})$ and $\mathcal{R}(\hat{z})$ denote the bitrates consumed by $\hat{y}$ and $\hat{z}$, respectively, $\mathcal{D}(x, \hat{x})$ measures the distortion between the input image $x$ and its reconstruction $\hat{x}$, $\lambda$ is a hyperparameter governing the rate-distortion trade-off, and $\lambda_z$ controls the bitrate allocation between the primary coding branch and the hyperprior branch.

\subsection{Analysis of Different Dimensions}\label{sec:exp of dim}

We conduct preliminary experiments to examine impacts of information loss across different dimensions on reconstruction quality. Specifically, using a non-autoregressive LIC model as the baseline, we simulate packet loss in latent space along either the spatial or channel dimension. Results are presented in Figure~\ref{fig:spatial and channel}. Packet loss in the spatial dimension results in masked pixel regions in the reconstruction, with PSNR degradation relatively uniform across images, albeit with a large absolute drop. In contrast, channel-dimension packet loss leads to highly inconsistent PSNR degradation. Some images experience negligible quality loss, while others suffer significant degradation. This discrepancy stems from the severely imbalanced energy distribution across channels in latent space. Loss of a high-energy channel, such as the 179th channel, causes complete decoding failure. Since packet loss is inherently stochastic, it is desirable to ensure that information carried by each packet is as balanced as possible. While spatial-dimension information is already relatively well-distributed, explicit design is required in the channel dimension to achieve comparable information balance.

\subsection{Inter-Channel Redistribution}
Directly mapping channels to packets is impractical, as loss of packets containing critical channels is unrecoverable. To address this challenge, we introduce an Inter-Channel Redistribution(ICR) module and its counterpart, Inverse Inter-Channel Redistribution (Inv-ICR), whose architectures are illustrated in Figure~\ref{fig:network overview}. At the encoder, the ICR module combines attention mechanisms, group rearrangement, and adaptive fusion to redistribute channel information across the latent representation $y$, achieving a more uniform energy distribution. At the decoder, the Inv-ICR module performs inverse operations, restoring original channel arrangement with channel unshuffle and channel-based attention. This implicit inter-channel interaction mechanism disperses packet energy, mitigating the impact of critical packet loss on decoding and enabling stable, high-quality reconstruction under diverse packet loss conditions.

\begin{table*}[t]
\centering
\caption{Performance comparison of models under packet loss with a uniform distribution on Kodak dataset.}
\label{tab:performance-comparison-kodak}
\small
\begin{tabular*}{\textwidth}{@{\extracolsep{\fill}} l l *{9}{c}}
\toprule
\multirow{2}{*}{$p_e$} & \multirow{2}{*}{Method} 
& \multicolumn{3}{c}{Low Bitrate} 
& \multicolumn{3}{c}{Medium Bitrate} 
& \multicolumn{3}{c}{High Bitrate} \\
\cmidrule(lr){3-5} \cmidrule(lr){6-8} \cmidrule(lr){9-11}
& 
& bpp & mean $\uparrow$ & var $\downarrow$ 
& bpp & mean $\uparrow$ & var $\downarrow$ 
& bpp & mean $\uparrow$ & var $\downarrow$ \\
\midrule
\multirow{7}{*}{5\%} 
& JPEG2000  & 0.165 & 24.116 & 0.117 & 0.216 & 24.556 & 0.201 & 0.347 & 25.278 & 0.243 \\
& ProgDTD & 0.133 & 26.312 & 0.436 & 0.221 & 26.877 & 0.820 & 0.359 & 27.919 & 0.564 \\
& LossResilientLIC &  0.131 &  26.850 &  0.015 &  0.210 &  28.247 &  0.043 &  0.338 &  29.561 &  0.007 \\
& ResiComp & 0.169 & 27.202 & 0.155 & 0.216 & 28.572 & 0.363 & 0.370 & \textbf{30.461} & 0.021 \\
& Baseline & 0.128 & 26.934 & 0.064 & 0.216 & 27.013 & 0.297 & 0.332 & 29.194 & 0.562 \\
& Random & 0.130 & 27.251 & 0.045 & 0.208 & 28.497 & 0.068 & 0.334 & 29.704 & 0.021 \\
& Ours &  0.125 &  \textbf{27.546} &  \textbf{0.002} & 0.217 &  \textbf{28.980} &  \textbf{0.001} & 0.346 &  30.433 &  \textbf{0.001}  \\
\addlinespace
\multirow{7}{*}{10\%} 
& JPEG2000  & 0.165 & 23.168 & 0.161 & 0.216 & 23.336 & 0.168 & 0.347 & 23.692 & 0.237 \\
& ProgDTD & 0.133 & 24.701 & 0.979 & 0.221 & 25.695 & 1.007 & 0.359 & 26.711 & 0.604 \\
& LossResilientLIC &  0.131 &  26.342 &  0.061 &  0.210 &  27.605 &  0.100 &  0.342 &  28.801 & 0.047 \\
& ResiComp & 0.169 & 26.991 & 0.178 & 0.216 & 28.516 & 0.141 & 0.370 & 30.232 & 0.077 \\
& Baseline & 0.128 & 25.946 & 0.498 & 0.216 & 26.489 & 0.188 & 0.332 & 27.385 & 0.712 \\
& Random & 0.130 & 26.819 & 0.099 & 0.208 & 28.103 & 0.077 & 0.334 & 29.131 & 0.114 \\
& Ours &  0.125 &  \textbf{27.470} &  \textbf{0.009} & 0.217 &  \textbf{28.845} &  \textbf{0.005} & 0.346 &  \textbf{30.271} &  \textbf{0.002}  \\
\addlinespace
\multirow{7}{*}{20\%} 
& JPEG2000  & 0.165 & 21.274 & 0.257 & 0.216 & 21.366 & 0.275 & 0.347 & 21.384 & 0.295 \\
& ProgDTD & 0.133 & 23.625 & 0.518 & 0.221 & 23.740 & 1.258 & 0.359 & 23.919 & 1.913 \\
& LossResilientLIC &  0.131 &  25.578 & 0.108 &  0.210 &  26.617 &  0.043 &  0.346 &  27.966 & 0.043 \\
& ResiComp & 0.169 & 24.046 & 0.563 & 0.216 & 27.390 & 0.234 & 0.370 & 29.540 & 0.077 \\
& Baseline & 0.128 & 23.534 & 2.011 & 0.216 & 22.724 & 1.344 & 0.332 & 23.753 & 2.034 \\
& Random & 0.130 & 26.482 & 0.107 & 0.208 & 27.441 & 0.187 & 0.334 & 28.168 & 0.114 \\
& Ours &  0.125 &  \textbf{27.211} &  \textbf{0.012} & 0.217 &  \textbf{28.561} & \textbf{0.010} & 0.346 &  \textbf{29.931} &  \textbf{0.003} \\
\bottomrule
\end{tabular*}
\end{table*}

\subsection{Interleaved Channel Grouping}\label{sec:ICG}

In practical transmission systems, commercial short-burst data protocols typically restrict packet sizes to 900 or 1500 bytes. Simply encapsulating compressed image features is impractical, as the resulting packet size would exceed limit even at low bitrates. To comply with constraints, we propose an Interleaved Channel Grouping (ICG) method to disperse information across packets. As shown in Figure~\ref{fig:network overview}, the latent $y$ is partitioned into two slices. For the first slice $s_1$, index-based partitioning is applied to derive $y_1$, comprising even-indexed channels, and $y_2$, comprising odd-indexed ones. Analogously, the second slice $s_2$ is partitioned to yield $y_3$ and $y_4$ in the same manner. For each $y_i$, the number of packets $N_i$ is determined based on its bitrate and the packet size constraint. The channel-dimension partition stride is set to $N_i$ accordingly, ensuring that $y_i$ is split into $N_i$ independent packets in an interleaved manner. With this partitioning strategy, all packets are endowed with comparable importance, consequently ensuring that loss of any specific packet does not lead to substantial degradations in reconstruction quality.

\subsection{Two-Layer Dual-Branch Autoregression}\label{sec:TDA}
The loss of a single packet results in the complete loss of all information it carries, potentially disrupting the decoding of other content that depends on it. However, a naive non-autoregressive approach comes with limitations of entropy modeling capacity. Therefore, we adopt a two-layer dual-branch autoregressive architecture. 

To mitigate cascading effects of autoregression, we establish short cross-slice dependencies, a two-layer design where decoding of $s_2$ is contingent upon $s_1$. As for our dual-branch design, an autoregressive relationship is established between $y_1$ and $y_3$, and between $y_2$ and $y_4$. Specifically, $y_1$ and $y_2$ are encoded first, and their learned probability models are leveraged to estimate distribution parameters of $y_3$ and $y_4$, facilitating entropy estimation under low-bitrate conditions. Despite the fact that a single packet loss within $y_1$ can impair probability estimation for $y_3$, under our proposed training strategy, information allocated to $y_3$ and $y_4$ is of low criticality, such that their loss has minimal impact on overall reconstruction quality.

\begin{table*}[t]
\centering
\caption{Gains over Baseline under packet loss with a uniform distribution on CLIC dataset.}
\label{tab:performance-comparison-clic}
\small
\begin{tabular*}{\textwidth}{@{\extracolsep{\fill}} l l *{9}{c}}
\toprule
\multirow{2}{*}{$p_e$} & \multirow{2}{*}{Method} 
& \multicolumn{3}{c}{Low Bitrate} 
& \multicolumn{3}{c}{Medium Bitrate} 
& \multicolumn{3}{c}{High Bitrate} \\
\cmidrule(lr){3-5} \cmidrule(lr){6-8} \cmidrule(lr){9-11}
& 
& bpp & mean $\uparrow$ & $\Delta$ $\uparrow$ 
& bpp & mean $\uparrow$ & $\Delta$ $\uparrow$ 
& bpp & mean $\uparrow$ & $\Delta$ $\uparrow$ \\
\midrule
\multirow{6}{*}{5\%}
& LossResilientLIC Baseline & 0.102 & 28.119 & - & 0.160 & 28.644 & - & 0.247 & 29.586 & - \\
& LossResilientLIC Random   & 0.102 & 28.379 & 0.260 & 0.164 & 29.737 & 1.093 & 0.258 & 30.958 & 1.372 \\
& LossResilientLIC          & 0.100 & 28.795  & 0.676 & 0.158 & 30.074 & 1.430 & 0.252 & 31.154 & 1.568 \\
& Our Baseline              & 0.090 & 29.197 & - & 0.147 & 30.299 & - & 0.230 & 31.078 & - \\
& Our Random                & 0.091   & 30.106 & 0.909 & 0.148 & 31.096 & 0.797 & 0.225 & 31.736 & 0.658 \\
& Ours                      & 0.090  & \textbf{30.550} & \textbf{1.353} 
                            & 0.148  & \textbf{31.928}  & \textbf{1.629} 
                            & 0.228  & \textbf{33.262} & \textbf{2.184} \\
\addlinespace
\multirow{6}{*}{10\%}
& LossResilientLIC Baseline & 0.102 & 26.450 & - & 0.160 & 27.290 & - & 0.247 & 27.638 & - \\
& LossResilientLIC Random   & 0.102 & 27.329 & 0.879 & 0.164 & 28.697 & 1.407 & 0.258 & 30.037 & 2.399 \\
& LossResilientLIC          & 0.100 & 28.261  & 1.811 & 0.158 & 29.210 & 1.920 & 0.252 & 30.491 & 2.853 \\

& Our Baseline              & 0.090 & 27.152 & - & 0.147 & 27.862 & - & 0.230 & 28.544 & - \\
& Our Random                & 0.091   & 29.300 & 2.148 & 0.148 & 30.004 & 2.142 & 0.225 & 30.276 & 1.732 \\
& Ours                      & 0.090  & \textbf{30.212} & \textbf{3.060} 
                            & 0.148  & \textbf{31.461}  & \textbf{3.599} 
                            & 0.228  & \textbf{32.990} & \textbf{4.446} \\
\addlinespace

\multirow{6}{*}{20\%}
& LossResilientLIC Baseline & 0.102 & 23.460 & - & 0.160 & 23.994 & - & 0.247 & 24.714 & - \\
& LossResilientLIC Random   & 0.104 & 26.781 & 3.321 & 0.174 & 26.957 & 2.963 & 0.265 & 28.879 & 4.165 \\
& LossResilientLIC          & 0.103 & 27.316  & 3.856 & 0.167 & 27.873 & 3.879 & 0.257 & 29.469 & 4.755 \\
& Our Baseline              & 0.090 & 24.056 & - & 0.147 & 24.702 & - & 0.230 & 25.818 & - \\
& Our Random                & 0.091   & 27.417 & 3.361 & 0.148 & 27.940 & 3.238 & 0.225 & 28.244 & 2.426 \\
& Ours                      & 0.090  & \textbf{29.862} & \textbf{5.806} 
                            & 0.148  & \textbf{31.200}  & \textbf{6.498} 
                            & 0.228  & \textbf{32.508} & \textbf{6.690} \\
\bottomrule
\end{tabular*}
\end{table*}

\subsection{Packet Loss Design}

During training, packet loss is simulated in the feature domain, whereas during testing, it is implemented by discarding corresponding binary files. The ICG module yields a set of packets that conform to the specified size constraints. Although LossResilientLIC~\cite{Sha_25_AAAI_LossResilientLIC} employs a packetization scheme based on spatial-channel rearrangement, it simulates packet loss at the individual channel level during training, which may result in scenarios where some channels within a packet are lost while others are retained. This is inconsistent with real-world transmission behavior. Given the inter-channel correlations within a packet, we adopt a packet-level loss simulation strategy instead. Specifically, for each lost packet, a binary mask $M^{C} \in \{0,1\}^{C}$ is applied to zero out all channels within the affected packet, where $C$ denotes the total number of channels. Based on the channel index set $I_k$ of the lost packet $k$, the corresponding mask entries are set to zero, i.e., $M_i = 0, \forall i \in I_k$. Furthermore, packet loss in the first-layer slice additionally triggers masking of the dependent entries in the second-layer slice in accordance with autoregressive dependencies. This packet-level loss simulation enables the model to handle real-world scenarios where $y_3$ and $y_4$ may experience degraded decoding due to inaccurate autoregressive estimation caused by upstream packet loss. Prior to decoding, the mask is fed into a mask-conditioned aggregation module~\cite{Sha_25_AAAI_LossResilientLIC}, which identifies the positions of lost channels and performs feature restoration.

\section{Experiment}
\subsection{Implementation Details}
\textbf{Datasets.} Our model is trained on the high-quality Flickr2W dataset \cite{Liu_2020_arXiv_Flickr}, with images randomly cropped to $256 \times 256$ pixels and a batch size of 8. For evaluation, we employ Kodak dataset~\cite{Franzen_93_url_kodak}, which contains images at a resolution of $768 \times 512$, and CLIC2020 dataset~\cite{Toderici_20_clic}, which contains images at 2K resolution. 

\noindent\textbf{Training Settings.} We adopt the LIC model HPCM~\cite{Li_25_ICCV_HPCM} into a non-autoregressive variant as our baseline. The training objective follows Equation~\ref{eq:loss}, with mean squared error (MSE) as the distortion metric to meet the high-fidelity requirements of low-bandwidth communication. The Lagrangian multiplier $\lambda$ is set to 0.0018, 0.0035, and 0.0067, corresponding to average bitrates of 0.125 bpp, 0.217 bpp, and 0.346 bpp on Kodak dataset, respectively.

The training procedure consists of three stages, each comprising $5\times10^5$ steps. In the first stage, modules related to packet loss are excluded, and training focuses exclusively on optimizing rate-distortion performance with a learning rate of $10^{-4}$. In the second stage, all modules are activated and random packet loss is introduced on the quantized latent representation $\hat{y}$. The maximum packet loss probability $p_{\text{max}}$ follows a stepwise schedule that progressively increases from 0\% to 30\% over training epochs, with each packet independently subject to random loss at probability $p \sim \mathcal{U}[0, p_{\text{max}}]$ within each batch. In the third stage, $p_{\text{max}}$ is fixed at 0.3, and the learning rate is decayed every $1.25\times10^5$ steps to fine-tune the model for optimal performance.

\noindent\textbf{Evaluation Setup.} Model evaluation incorporates the full packet encapsulation process. Following commercial short-burst data specifications, we impose a packet size constraint of 1500~bytes. Under our ICG scheme, the hyperprior feature $z$ is encapsulated as a standalone packet, while each $y_i$ is partitioned into $N_i$ independent packets along the channel dimension. All packets are stored as binary files via arithmetic coding. The number of packet partitions varies across images depending on the bitrate, with detailed packaging configurations described in Section~\ref{sec:ICG}. During decoding, certain files are randomly discarded to emulate packet loss in real-world transmission. Two packet loss simulation protocols are considered, uniform loss and trace-driven loss. Uniform loss rates are set at 5\%, 10\%, and 20\%. Note that the hyperprior bitstream must be reliably received, as it is essential for estimating the latent distribution in the LIC hyperprior module. In our scheme, the hyperprior bitstream accounts for less than 8\% of the total bitstream and can be encapsulated within a single packet.  All experiments are conducted on a single GPU with 24 GB VRAM.

\begin{figure*}[t]
  \includegraphics[width=\textwidth]{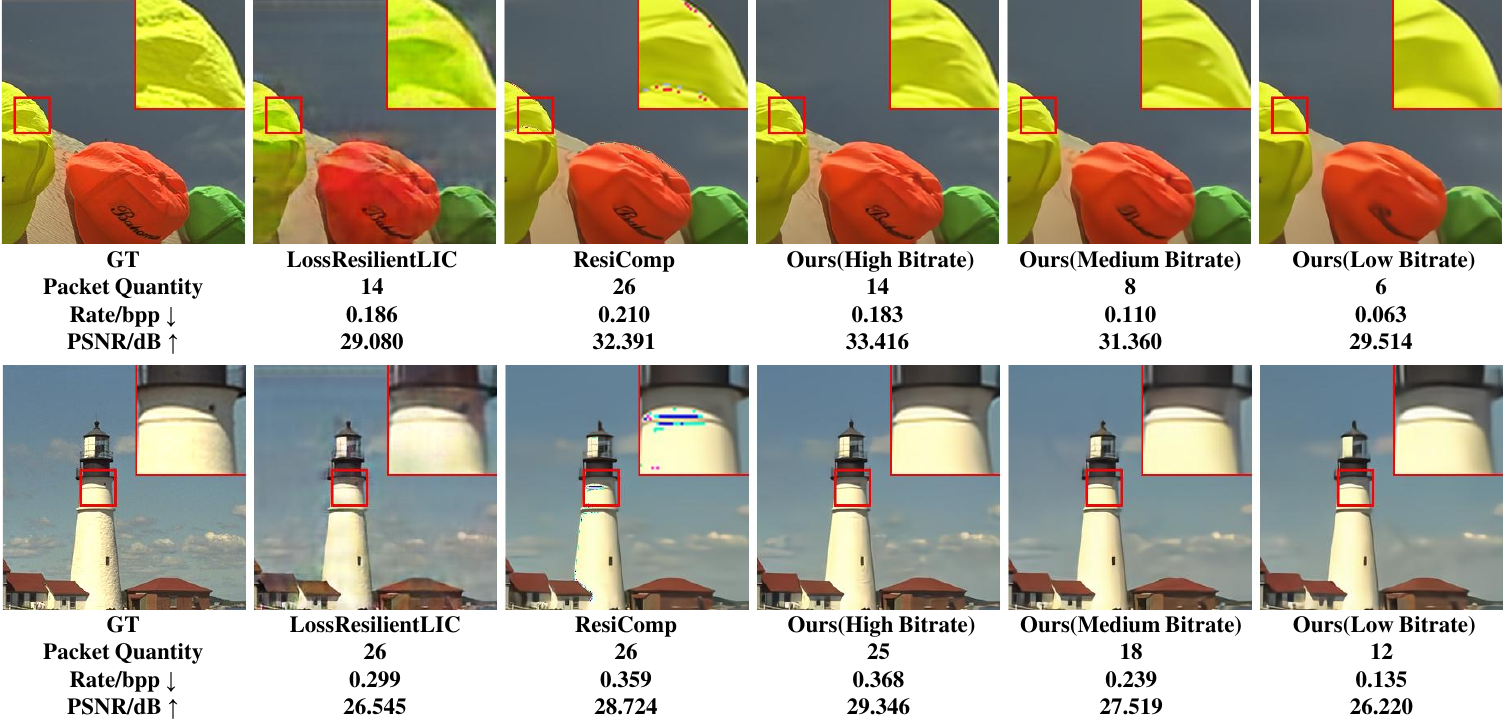}
    \caption{Visualization of reconstructed images on Kodak dataset. All methods lose the first two packets. Our proposed method achieves better results at a lower bitrate.}
  \Description{fig:vision}
  \label{fig:vision}
\end{figure*}

\subsection{Performance Under Uniform Loss}
We begin by presenting a comparative analysis of the rate-distortion performance under uniform packet loss. In our experiments, "Baseline" refers to a non-autoregressive variant based on HPCM~\cite{Li_25_ICCV_HPCM}, with implementation details specified in the supplementary material. The random mask model, denoted as "Random", extends "Baseline" by incorporating channel-wise random packet loss and the MCA module~\cite{Sha_25_AAAI_LossResilientLIC}. For comprehensive comparisons, we additionally incorporate experimental results from JPEG2000~\cite{Skodras_01_jpeg2000}, ProgDTD~\cite{Hojjat_23_CVPR_ProgDTD}, LossResilientLIC~\cite{Sha_25_AAAI_LossResilientLIC}, and ResiComp~\cite{Wang_25_TCSVT_ResiComp}. JPEG2000 adopts a layered progressive transmission scheme, and we assume base layers remain loss-free for fair. For ProgDTD, we follow its source code and evaluate it under uniform packet loss conditions. Results for LossResilientLIC are sourced from the original paper. For ResiComp, we use Layered and LayeredMDC modes at low bitrates, and Intra Slice mode at higher bitrates. Analogous to JPEG2000, base slices are assumed to be loss-free. To account for the stochasticity of packet loss, all results are averaged over 10 independent trials, with variance reported to characterize reconstruction stability.

Experimental results on Kodak dataset are presented in Table~\ref{tab:performance-comparison-kodak}. Our method achieves state-of-the-art reconstruction quality across virtually all bitrate regimes and packet loss rates, while simultaneously attaining the lowest variance in PSNR across all evaluated loss conditions. Specifically, at a moderate packet loss rate of $p_e = 10\%$, our model outperforms ResiComp by 0.48~dB, 0.33~dB, and 0.73~dB at low, medium, and high bitrates, respectively, while consuming much less bandwidth. As $p_e$ increases to 20\%, the advantage of our method becomes even more pronounced, with PSNR gains of 3.17~dB and 1.17~dB over ResiComp at low and medium bitrates, demonstrating significantly superior robustness under severe loss conditions. The variance metric further highlights the stability of our method. While competing approaches exhibit variance values exceeding 0.1 at $p_e = 20\%$, our model consistently maintains variance below 0.012, which is an order of magnitude lower than LossResilientLIC and ResiComp. This stability is critical in practice, as it implies that reconstruction quality remains reliably high regardless of the specific pattern of packet loss. Although our PSNR at $p_e = 5\%$ high bitrate is marginally below that of ResiComp, our method achieves this at a 6.5\% lower bitrate and with two orders of magnitude lower variance, making it clearly more effective in bandwidth-constrained and reliability-sensitive scenarios.

\subsection{Comparison via Gain}
Table~\ref{tab:performance-comparison-clic} presents results on CLIC dataset, where our model is compared against LossResilientLIC and its intermediate variants. In addition to absolute PSNR, we report $\Delta$, the gain in mean PSNR over respective baselines, which isolates the loss-resilience gain from the codec baseline. Our method consistently achieves the highest absolute PSNR and the largest $\Delta$ across all settings. Our model yields more gains compared to LossResilientLIC, and the advantage grows substantially under heavier packet loss. At $p_e = 20\%$, our method consistently outperforms LossResilientLIC, achieving improvements of over 1.8~dB across all bitrates. These results confirm that our model possesses stronger and more scalable loss-resilience than competing methods. As the channel degrades, our model increasingly outperforms LossResilientLIC in absolute quality.

Taken together, the results on both datasets demonstrate that our proposed method achieves a superior balance of rate-distortion performance, anti-packet-loss robustness, and prediction stability.

\begin{figure}[t]
  \centering
  \includegraphics[width=\linewidth]{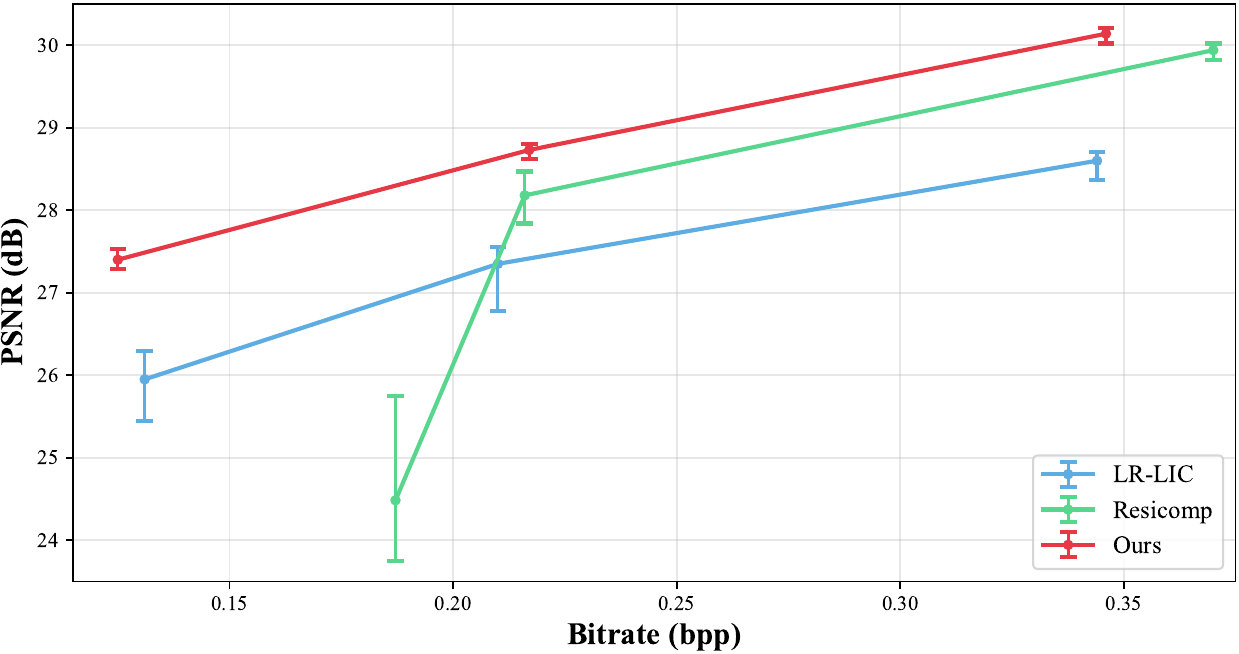}
  \caption{Rate-distortion performance under GE model simulation. Each data point reports the mean PSNR, and the vertical extent of the error bar indicates the variance across 10 independent trials, reflecting both reconstruction quality and stability under bursty packet loss.}
  \label{fig:ge result}
  \Description{fig:ge result}
\end{figure}

\begin{figure*}[t]
  \includegraphics[width=\textwidth]{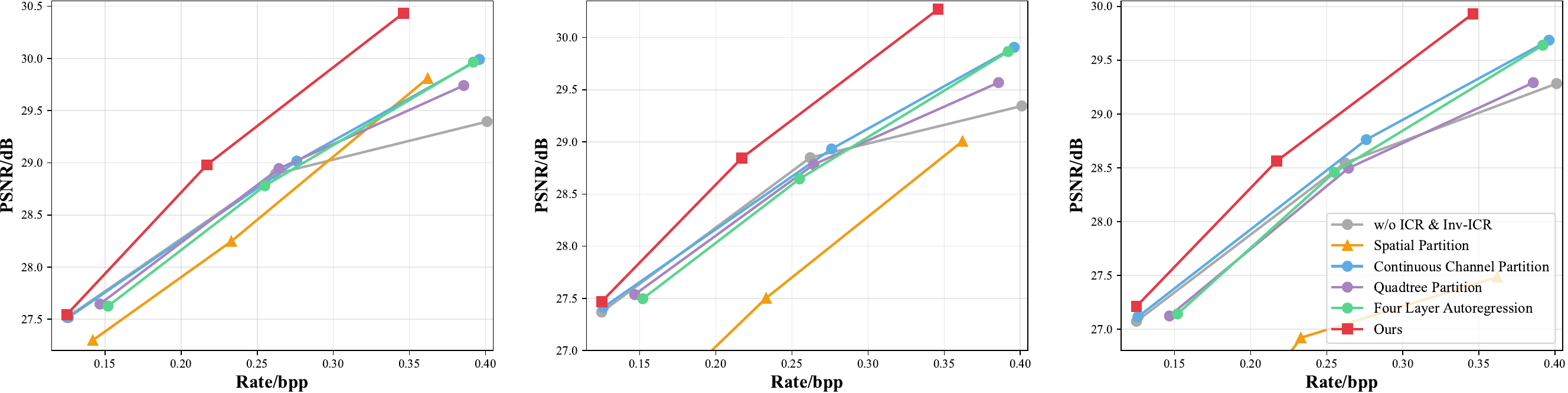}
  \caption{Ablation on module design, partition strategy, and autoregression mode at packet loss rates of 5\%, 10\%, and 20\% (left to right). \textit{Ours} adopts ICR and Inv-ICR modules with interleaved channel grouping and two-layer dual-branch autoregression.}
  \Description{fig:ablation}
  \label{fig:ablation}
\end{figure*}

\subsection{Performance Under Simulated Loss}
To further assess the generalization of our scheme to realistic network conditions, we evaluate it under the Gilbert-Elliott (GE) model~\cite{Hasslinger_08_MME_gilbert}. Real-world transmission environments are often characterized by complexity and variability, where fluctuations in long-distance communication give rise to bursty packet losses that cannot be adequately captured by uniform loss models. The GE model addresses this limitation through a two-state Markov chain that simulates communication channels switching between Good and Bad states with certain probabilities, and has been widely adopted for modeling channel errors and evaluating loss-resilient strategies.

The steady-state probabilities of Good and Bad states are given by $\frac{r}{p+r}$ and $\frac{p}{p+r}$, respectively. Denoting the loss rates in Good and Bad states as $l_g$ and $l_b$, the overall theoretical packet loss rate is expressed as $P_E = \frac{r}{p+r} l_g + \frac{p}{p+r} l_b$. A key characteristic of this model is that the state at any given time influences the subsequent state; following a packet loss event, the probability of losing the next packet is elevated, faithfully reflecting the bursty nature of real network losses. This temporal dependency provides a more rigorous basis for evaluating loss resilience than memoryless uniform loss.

We instantiate the GE model with parameter set $[p, r, l_g, l_b] = [0.417, 0.973, 0.052, 0.380]$, corresponding to an average packet loss rate of approximately 15\%, and evaluate LossResilientLIC, ResiComp, and our method. As depicted in Figure~\ref{fig:ge result}, our model consistently outperforms competing methods. It achieves the highest mean PSNR at every operating point, with the margin becoming more pronounced at lower bitrates where robustness to bursty loss is most critical. Error bars of our method are substantially narrower than those of competing approaches, indicating significantly lower variance across independent trials. This is particularly evident for ResiComp, whose large error bars at medium bitrates reveal considerable sensitivity to specific packet loss. These findings are consistent with the conclusions drawn from the uniform loss experiments, confirming that the superiority of our method generalizes across different loss models. Notably, while LossResilientLIC explicitly incorporates a GE channel model during training, our model is trained exclusively under uniform loss. Dispite no GE-base simulation during training, our model achieves superior performance, which we attribute to well-balanced information distributions across packets facilitated by our proposed ICR and ICG mechanism.

\subsection{Visual Results}
To provide a more intuitive illustration of our method's performance under packet loss, we conduct a visual comparison of first-two-packet loss among LossResilientLIC, ResiComp, and our approach. For LossResilientLIC, the front packets encapsulate more critical information owing to its drop-tail designs. Consequently, if the first two packets are lost, the reconstruction quality remains poor even when all subsequent packets are successfully received, and reconstructed images exhibit gray artifacts in regions corresponding to lost packets. A similar limitation is observed in Layered mode of ResiComp, where packet loss leads to visible decoding failures manifested as corrupted object boundaries in the reconstructed image. Although Intra Slice mode offers better anti-packet-loss capabilities, its rate-distortion performance is suboptimal. In contrast, our proposed method is insensitive to the position of loss, producing visually clean reconstructions free of artifacts or boundary distortions, while achieving superior PSNR at lower bitrates.

\subsection{Ablation Study}

We conduct ablation experiments to validate the contribution of each proposed component, including the ICR mechanism, the ICG strategy, and the two-layer dual-branch autoregression design. Results are reported across three packet loss rates in Figure~\ref{fig:ablation}, with quantitative results and complexity analysis shown in Appendix.

Removing ICR and Inv-ICR modules leads to significant performance degradation, especially at high bitrates and loss rates, confirming that these modules are critical for achieving robust reconstruction under packet loss. Replacing the ICG module with other partition strategies results in a consistent performance drop, demonstrating that interleaved grouping is essential for alleviating non-uniform distribution of information across packets and thereby improving loss resilience. Notably, the spatial partitioning strategy is the least effective for packetization, which is consistent with analysis in Section~\ref{sec:exp of dim}. Increasing the number of autoregressive layers from two to four yields degradation that increases with bitrate, suggesting that deeper autoregression introduces a longer dependency chain without contributing to robustness. The full model combining all proposed components achieves the best performance across all conditions, and the performance gap relative to the ablated variants widens as the packet loss rate increases, further underscoring the importance of each design choice.

\section{Conclusion and Limitation}

In this paper, we propose a loss-resilient image compression scheme based on channel-wise packetization and hierarchical prediction. By introducing the Inter-Channel Redistribution mechanism alongside the Interleaved Channel Grouping strategy, our method achieves an even information dispersal across packets. Extensive experiments demonstrate that our scheme consistently outperforms existing approaches in terms of both reconstruction quality and stability across varying packet loss rates and transmission environments.

Our method shares a common limitation with all hyperprior-based LIC approaches that the hyperprior bitstream must be received intact for correct decoding. In practice, protecting hyperprior with FEC, such as Reed–Solomon coding~\cite{Reed_1960_SIAM_polynomial}, incurs only about 7\% bandwidth overhead while achieving over 99\% recovery probability under GE-based simulations. We also leave the exploration of deep learning‑based protection for the hyperprior as future work.

\begin{acks}
This work was supported by the National Key Research and Development Program of China under Grant 2024YFF0509700, the National Natural Science Foundation of China (62471290, 62431015, 62331014) and the Fundamental Research Funds for the Central Universities.
\end{acks}

\bibliographystyle{ACM-Reference-Format}

\bibliography{main}

\newpage
\appendix

\section{Details of Architecture}

\begin{figure*}[t]
  \includegraphics[width=\linewidth]{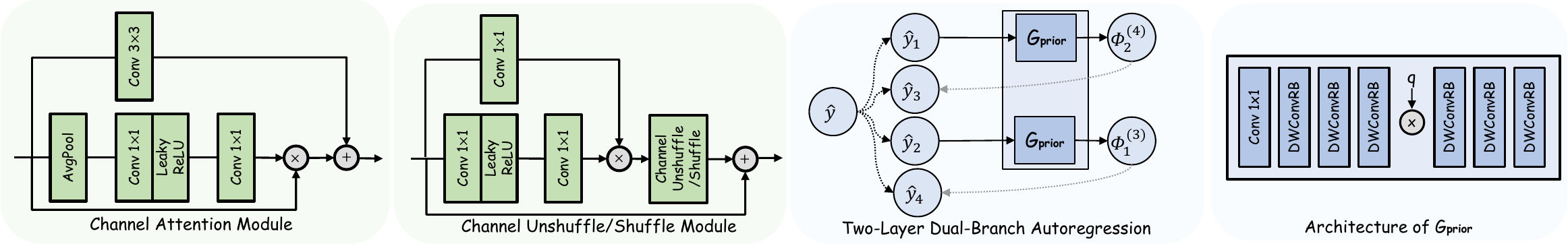}
  \caption{The left two blocks depict details of Channel Unshuffle/Shuffle and Channel Attention modules in ICR and Inv-ICR. The right two blocks represent the dependency chain of the dual-Branch autoregressive structure and details of \(g_{prior}\), respectively. The coding of \(y_3\) is independent of \(y_2\), and likewise, that of \(y_4\) is independent of \(y_1\).}
  \label{fig:appendix archi auto}
  \Description{fig:appendix archi auto}
\end{figure*}

We adopt the state-of-the-art learned image compression framework, HPCM, with only the hierarchical coding schedule at multiple scales being modified. As the Baseline model, we employ a non-autoregressive architecture with independent estimation of each latent element, serving as a performance reference in the absence of packet-loss-aware training. The “Random” model introduces a random masking strategy and the Mask Conditional Aggregation (MCA) module, establishing a performance benchmark for packet-loss-aware training.

Building upon the “Random” model, our proposed method further incorporates three key design components, including an Inter-Channel Redistribution (ICR) mechanism, an Interleaved Channel Grouping (ICG) packetization design, and a two-layer dual-branch autoregressive structure. The architecture of ICR, Inv-ICR, and ICG modules have been illustrated in main paper. The ICR mechanism redistributes high-energy channels, i.e., channel 179, across the latent feature by channel-wise attention and shuffle/unshuffle modules, the architectures of which are illustrated in Figure~\ref{fig:appendix archi auto}. As a result, all channels are reweighted to a state of comparable importance. We also provide a conceptual illustration of our two-layer dual-branch autoregression in Figure~\ref{fig:appendix archi auto}. In contrast to the limited entropy modeling capacity of non-autoregressive architectures, our two-layer autoregression encodes the first-slice elements to serve as contexts for predicting the second-slice ones, so that subsequent packets can accommodate more channels at a lower bit rate. The dual-branch design mitigates cascading errors from lost packets by confining loss impact to a single branch without cross-branch contamination. This also better aligns with the requirements of anti-packet-loss transmission under real-world scenarios.

\begin{table*}[t]
\centering
\caption{Comparison of protected payload sizes.}
\label{tab:protected_payload}
\begin{tabular*}{\linewidth}{l @{\extracolsep{\fill}} c c c c c c c c c c c c}
\toprule
\multirow{2}{*}{Measure} & \multicolumn{3}{c}{JPEG2000} & \multicolumn{3}{c}{ResiComp} & \multicolumn{3}{c}{LossResilientLIC} & \multicolumn{3}{c}{Ours} \\
\cmidrule(lr){2-4} \cmidrule(lr){5-7} \cmidrule(lr){8-10} \cmidrule(lr){11-13}
                  & low & medium & high & low & medium & high & low & medium & high & low & medium & high \\
\midrule
Total Bitrate     & 0.165 & 0.216 & 0.347 & 0.169 & 0.216 & 0.370 & 0.131 & 0.210 & 0.346 & 0.125 & 0.217 & 0.346 \\
Protected Payload & 775.8 & 775.8 & 775.8 & 1276.7 & 1236.5 & 1405.5 & $\sim 386.3$ & $\sim 619.3$ & $\sim 1020.4$ & 447.3 & 516.1 & 638.5 \\
Proportion        & 9.57\% & 7.31\% & 4.55\% & 15.37\% & 11.65\% & 7.73\% & $\sim 6\%$ & $\sim 6\%$ & $\sim 6\%$ & 7.23\% & 4.84\% & 3.76\% \\
\bottomrule
\end{tabular*}
\end{table*}

\section{Assumptions for Transmission}
Our proposed scheme features a hierarchical prior model that captures scale priors of latent elements via hyperprior information, to enhance the accuracy of distribution estimation and improve rate-distortion performance. However, loss of the hyperprior bitstream during transmission leads to biased latent distribution estimation, rendering subsequent packets undecodable. Therefore, reliable transmission of this bitstream is critical. 

\begin{figure*}[t]
  \includegraphics[width=\textwidth]{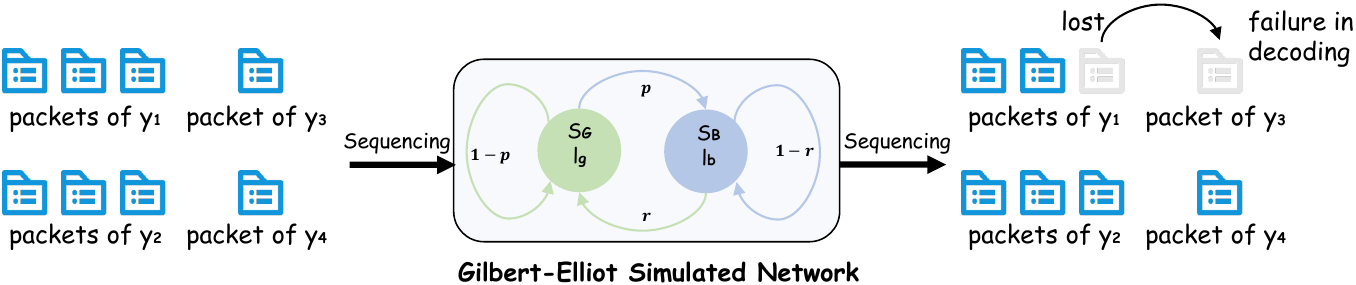}
  \caption{GE-based simulation of packet loss transmission.} 
  \Description{fig:appendix ge simulation}
  \label{fig:appendix ge simulation}
\end{figure*}

To address this challenge, we decompose the rate loss into prior and hyperprior components and triple the weight of hyperprior rate loss. With this design, the hyperprior bitstream accounts for 7.23\%, 4.84\%, and 3.76\% of the overall at low, medium, and high bitrates, respectively, based on our experimental results. 
Furthermore, the maximum size of the hyperprior bitstream does not exceed 750 bytes, satisfying the packet-size constraints of commercial short-burst data protocols. Therefore, the hyperprior can be protected with limited overhead using reliability enhancement techniques such as selective retransmission or forward error correction (FEC). A detailed analysis of the specific FEC configuration and protection overhead is provided in the following section.

It is worth noting that the requirement of protecting essential side information is not unique to our approach. Existing robust compression schemes also depend on a protected core payload for successful reconstruction. Specifically, JPEG2000 requires protection of the initial base layers, while ResiComp assumes the first two packets to be loss-free. Table~\ref{tab:protected_payload} compares the sizes of the protected payloads among different methods. As shown, the hyperprior in our approach occupies only a small fraction of the overall bitstream and is comparable to or smaller than the protected payloads required by competing schemes. This demonstrates that our reliability assumption is consistent with existing approaches and that the additional protection cost introduced by our method remains limited.

\section{FEC Cost for Hyperprior}
Under the 900-byte packetization constraint, the hyperprior bitstream can typically be transmitted within a single packet. However, since packet losses occur at the packet level, applying RS coding~\cite{Reed_1960_SIAM_polynomial} directly to a single packet cannot recover from a complete packet erasure. Therefore, when FEC protection is adopted, the hyperprior is partitioned into multiple packets to enable cross-packet redundancy and recovery.

Under the Gilbert--Elliott (GE) model with parameters $p=0.417$, $r=0.973$, $l_g=0.052$, and $l_b=0.380$, average packet loss rate is approximately 15\%, while the probability of experiencing five or more consecutive packet losses is below $10^{-6}$. Accordingly, we set the maximum burst loss length to $L=4$ and employ RS(12, 8) coding with a 64-byte packet format, including a 2-byte header and up to 62-byte payload. 

For a 452-byte hyperprior bitstream (e.g., Kodim11), the proposed configuration requires 8 source packets and 4 parity packets, resulting in a total transmission size of 768 bytes. As a result, the effective bitrate increases from 0.0898 bpp to 0.0962 bpp, corresponding to an additional overhead of approximately 7\%. Under the GE model, this RS configuration guarantees successful recovery whenever the number of lost packets does not exceed the four available redundancy packets, which occurs with a probability exceeding 99\%.

\section{Uniform Loss and Trace-Driven Loss}
We adopt two packet loss simulation protocols, uniform loss and trace-driven loss, to evaluate our method across different loss rates.

Due to varying packet counts across different images, multiplying the packet count directly by the loss rate often yields a non-integer number of lost packets, which introduces inaccuracies in loss rate simulation. To address this, uniform loss independently generates a random number for each packet based on the given loss rate to determine whether it is lost. Furthermore, to mitigate the effects of random fluctuations, we repeat each experiment ten times per configuration and report the averaged results.

For trace-driven loss, we adopt the Gilbert–Elliott (GE) model. Built upon a two-state Markov chain, the GE model defines state transition parameters \(p\) and \(r\), representing the probabilities of switching from the good state to the bad state and vice versa, along with state-dependent loss rates \(l_g\) and \(l_b\). The overall packet loss rate is given by \(P_E = \frac{r}{p+r} l_g + \frac{p}{p+r} l_b\). Unlike uniform loss, GE-based simulations propagate the current channel state to the next state, thereby better reflecting real-world transmission channel characteristics.

Under both above packet loss patterns, our method consistently achieves higher reconstruction quality with lower variance, underscoring its superiority and strong versatility across bandwidths and datasets.

\section{Additional Experimental Results}
\subsection{Error-Free Performance and Component Effects.}
\begin{figure}[t]
  \centering
  \includegraphics[width=\linewidth]{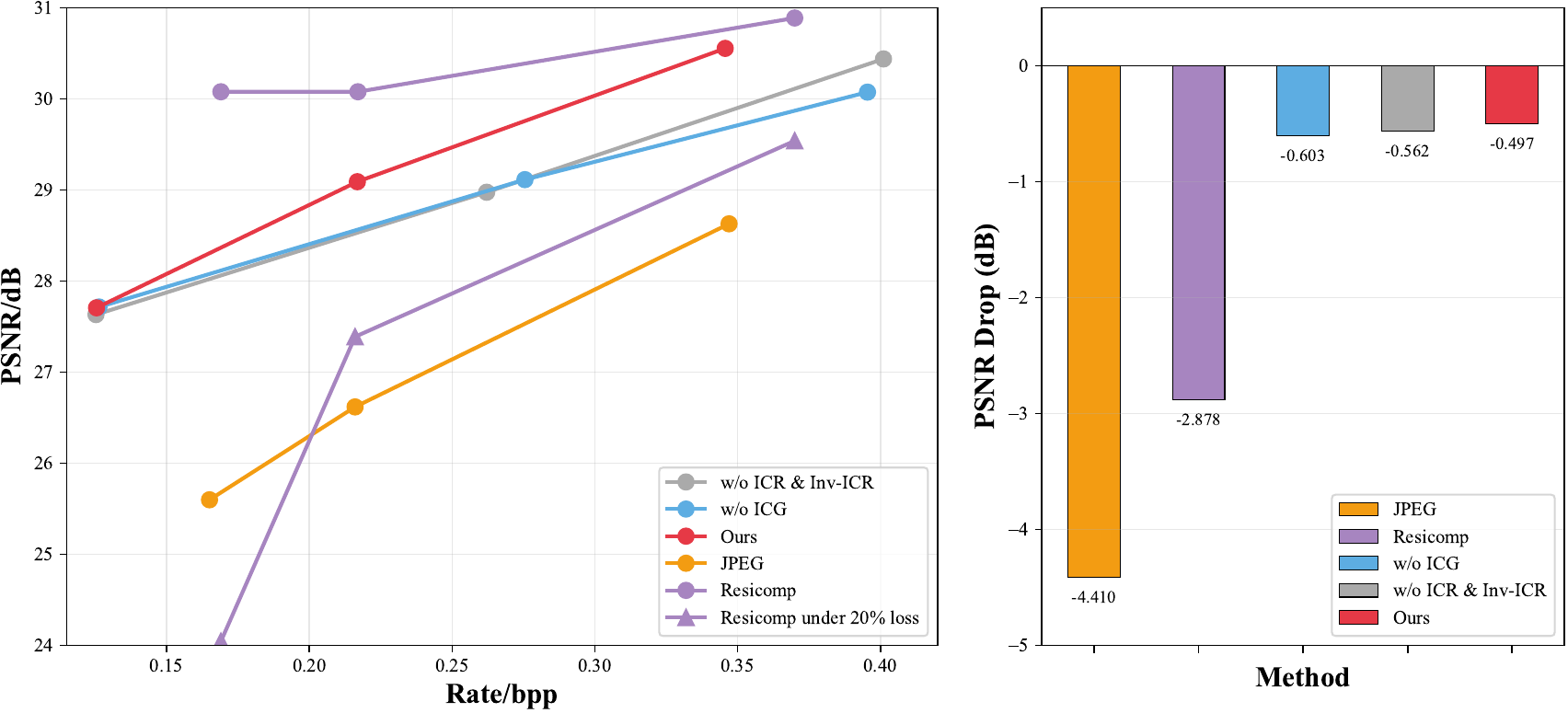}
  \caption{Rate-Distortion curves under no packet loss, and comparison of performance degradation under a 20\% packet loss rate.}
  \label{fig:rebuttal_noloss}
  \Description{fig:rebuttal_noloss}
\end{figure}

First, we provide a performance comparison of all schemes under the loss-free condition, along with qualitative degradations under 20\% packet loss rate in Figure~\ref{fig:rebuttal_noloss}. This demonstrates that the advantage of our method lies not only in its strong baseline performance, but also in its superior robustness against packet losses. 

ICR and Inv-ICR modules incur only a marginal RD penalty ($\approx$ 0.1 dB), while the ICG module and dual-branch autoregression improve compression performance. To verify that the advantage under packet loss does not simply stem from stronger lossless performance, we compare the PSNR degradation of different methods under packet loss. Our method consistently exhibits the smallest degradation, indicating that its robustness arises from the proposed designs rather than a more favorable error-free baseline.

\begin{table}[t]
\centering
\caption{BD-PSNR under different packet loss rates (dB).}
\label{tab:bd_psnr_full}
\begin{tabular*}{\columnwidth}{@{\extracolsep{\fill}} l c c c @{}}
\toprule
\textbf{Packet Loss Rate} & \textbf{5\%} & \textbf{10\%} & \textbf{20\%} \\
\midrule
Ours & 0.000 & 0.000 & 0.000 \\
w/o ICR Inv-ICR & -0.425 & -0.369 & -0.407 \\
w/o ICG & -0.500 & -0.462 & -0.372 \\
Quadtree Partition & -0.536 & -0.548 & -0.575 \\
Four Layer Autoregression & -0.627 & -0.623 & -0.556 \\
Spatial Partition & -0.849 & -1.530 & -2.019 \\
w/o Training Stage 1 & -0.599 & -0.558 & -0.495 \\
w/o Training Stage 2 & -0.885 & -0.809 & -0.680 \\
w/o Mask Propagation & -0.035 & -0.287 & -0.674 \\
\bottomrule
\end{tabular*}
\end{table}

\subsection{Comprehensive Results on Datasets}
The analysis of rate-distortion performances on the Kodak dataset and PSNR gains over the baseline model on CLIC dataset have been presented in main paper. For a more comprehensive comparison, we provide PSNR gains over the baseline on the Kodak dataset, as well as complete comparisons on the CLIC dataset. Experimental results demonstrate that our method achieves higher reconstruction quality and stronger stability at the same or even lower bitrates, consistent with the findings in main paper. 

Notably, to ensure fair comparisons with LossResilientLIC, we followed its setup and set the packet size to 4500 bytes on the CLIC dataset. The corresponding results for 900-byte packetization are also included. Due to the progressive coding of JPEG2000 and the concentration of critical information in preceding channels of LossResilientLIC, neither method supports 900-byte packetization. 

Layered and LayeredMDC modes of ResiComp involve decoding dependency chains, where packet loss causes error propagation. Consequently, with a 900-byte packet size, the increased number of encapsulated packets leads to uncontrollable degradations in reconstruction quality. In contrast, our method achieves excellent performance across different packet size configurations.

\begin{figure}[t]
  \centering
  \includegraphics[width=\linewidth]{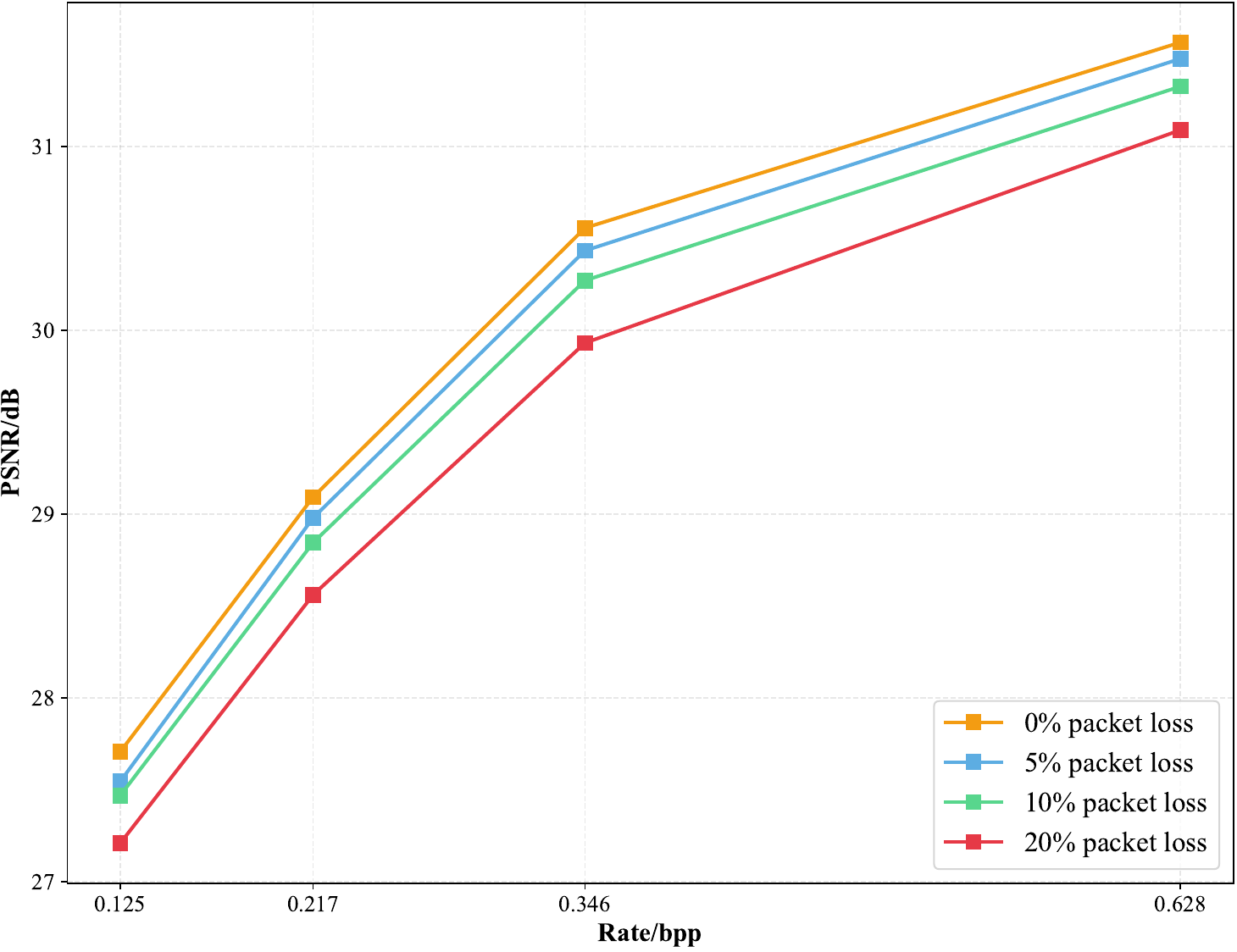}
  \caption{The rate-distortion curves of our model on the Kodak dataset across packet loss rates.}
  \label{fig:appendix rd}
  \Description{fig:appendix rd}
\end{figure}

\begin{table}[t]
\centering
\caption{Performance comparison of our method under specific packet loss on Kodim11 at 0.117~bpp. \( \times\) denotes that the packet in the corresponding column is lost.}
\label{tab:specific-loss-performance}
\small
\begin{tabular}{c c c c c c c c c c}
\toprule
\multicolumn{3}{c}{$y_1$} & \multicolumn{3}{c}{$y_2$} & \multicolumn{1}{c}{$y_3$} & \multicolumn{1}{c}{$y_4$} & \multirow{2}{*}{PSNR$\uparrow$} & \multirow{2}{*}{MS-SSIM$\uparrow$} \\
\cmidrule(lr){1-3} \cmidrule(lr){4-6} \cmidrule(lr){7-7} \cmidrule(lr){8-8}
$P_1$ & $P_2$ & $P_3$ & $P_4$ & $P_5$ & $P_6$ & $P_7$ & $P_8$ & & \\
\midrule
&   &   &   &   &   &   &   & 27.151 & 9.589 \\
\(\times\) &   &   &   &   &   & \(\times\)  &   & 26.970 & 9.484 \\
  & \(\times\) &   &   &   &   &  \(\times\) &   & 26.708 & 9.210 \\
  &   & \(\times\) &   &   &   & \(\times\)  &   & 26.807 & 9.442 \\
  &   &   & \(\times\) &   &   &   &  \(\times\) & 26.790 & 9.433 \\
  &   &   &   & \(\times\) &   &   &  \(\times\) & 26.916 & 9.375 \\
  &   &   &   &   & \(\times\) &   &  \(\times\) & 26.892 & 9.366 \\
\(\times\) & \(\times\) & \(\times\) &   &   &   & \(\times\)  &   & 25.652 & 8.701 \\
  &   &   & \(\times\) & \(\times\) & \(\times\) &   & \(\times\)  & 25.693 & 8.490 \\
\bottomrule
\end{tabular}
\end{table}

\subsection{Performance under Targeted Packet Loss}
We conduct targeted packet loss tests on the Kodak dataset, with results presented in Table~\ref{tab:specific-loss-performance}. For clearer comparisons, we map MS-SSIM to a distortion measure using \(-10 \log_{10}(1 - \operatorname{MS-SSIM})\). Regardless of which single packet is lost, our method maintains robust overall reconstruction quality without substantial degradations. Even with complete loss of the first or second slice, the PSNR drop is limited to within 1.5~dB.

\subsection{Ablation on Design}
The proposed architecture involves several key design choices, including the two-slice latent decomposition, the Interleaved Channel Grouping (ICG) strategy, the ICR/Inv-ICR modules, the three-stage training strategy, and masking propagation. We conduct comprehensive ablation studies to evaluate the contribution of each component to compression performance and loss resilience.

We use BD-PSNR in Table~\ref{tab:bd_psnr_full} to quantitatively measure the performance degradation caused by replacing or removing each proposed component. The results consistently validate the effectiveness of all proposed designs. Specifically, removing the ICR/Inv-ICR modules or replacing ICG with alternative partitioning strategies leads to noticeable performance degradation, demonstrating their importance for robust reconstruction under packet loss. Moreover, increasing the autoregressive depth does not improve robustness and instead introduces additional dependency, resulting in degraded performance under challenging loss conditions. The complete model achieves the best overall performance across different packet loss rates, with particularly significant advantages at higher loss rates.

We also provide additional ablations on the training strategy and masking propagation. Specifically, removing either the first or second training stage leads to performance degradation, while the gains from masking propagation become more pronounced as the packet loss rate increases.

It is worth noting that the proposed Two-Layer Dual-Branch Autoregression differs from the channel-wise autoregression strategy in~\cite{Minnen_20_ICIP_CC}. Specifically, \cite{Minnen_20_ICIP_CC} relies on multi-layer contextual dependencies for entropy prediction, leading to a longer dependency chain. This design is analogous to the deeper autoregressive configuration evaluated in our ablation study. Our results show that increasing autoregressive depth does not improve robustness and may even degrade performance under packet loss, especially at higher bitrates. This observation motivates the adoption of the proposed two-layer dual-branch design.

\begin{table}[t]
\centering
\caption{Complexity comparison on Kodak under 20\% packet loss.}
\label{tab:complexity}
\resizebox{\linewidth}{!}{
\begin{tabular}{l c c c c}
\toprule
\textbf{Model} & 
{\textbf{Params (M)}} & 
{\textbf{FLOPs (G)}} &  
{\textbf{Inf. Time (ms)}} &
{\textbf{BD-Rate}} \\
\midrule
ResiComp & 130.82 & 985.586 & 451 & 0 \\
Spatial Partition & 89.12 & 228.691 & 364 & -0.222\% \\
w/o ICG & 89.12 & 228.691 & 355 & -15.227\% \\
Four Layer Autoregression & 109.16 & 259.430 & 386 & -8.706\% \\
Quadtree Partition & 89.12 & 228.691 & 371 & -8.793\% \\
w/o ICR \& Inv-ICR & 84.33 & 221.386 & 352 & -14.534\% \\
Ours               & 89.12 & 228.691 & 355 & -25.274\% \\
\bottomrule
\end{tabular}
}
\end{table}

\section{Analysis of Complexity}
The ablation study demonstrates that the ICR mechanism provides a substantial PSNR improvement under packet loss. To further analyze its computational impact, we report the complexity of the ICR and Inv-ICR modules, including their parameter counts, FLOPs, and the resulting inference time increase, in Table~\ref{tab:complexity}. The results show that these modules introduce only negligible computational overhead.

Moreover, the proposed dual-branch autoregressive architecture does not introduce additional computational complexity compared with the commonly used quadtree partitioning strategy, while achieving lower inference time than conventional four-layer autoregressive entropy models. The proposed Interleaved Channel Grouping (ICG) replaces contiguous channel partitioning with stride-based grouping and is implemented with the same channel-wise partitioning operation, resulting in no additional computational cost. Overall, compared with ResiComp, our method reduces FLOPs by 77\% and achieves a 21\% reduction in inference time, demonstrating its superior efficiency.

\section{More Visual Results}
\begin{figure*}[t]
  \includegraphics[width=0.93\textwidth]{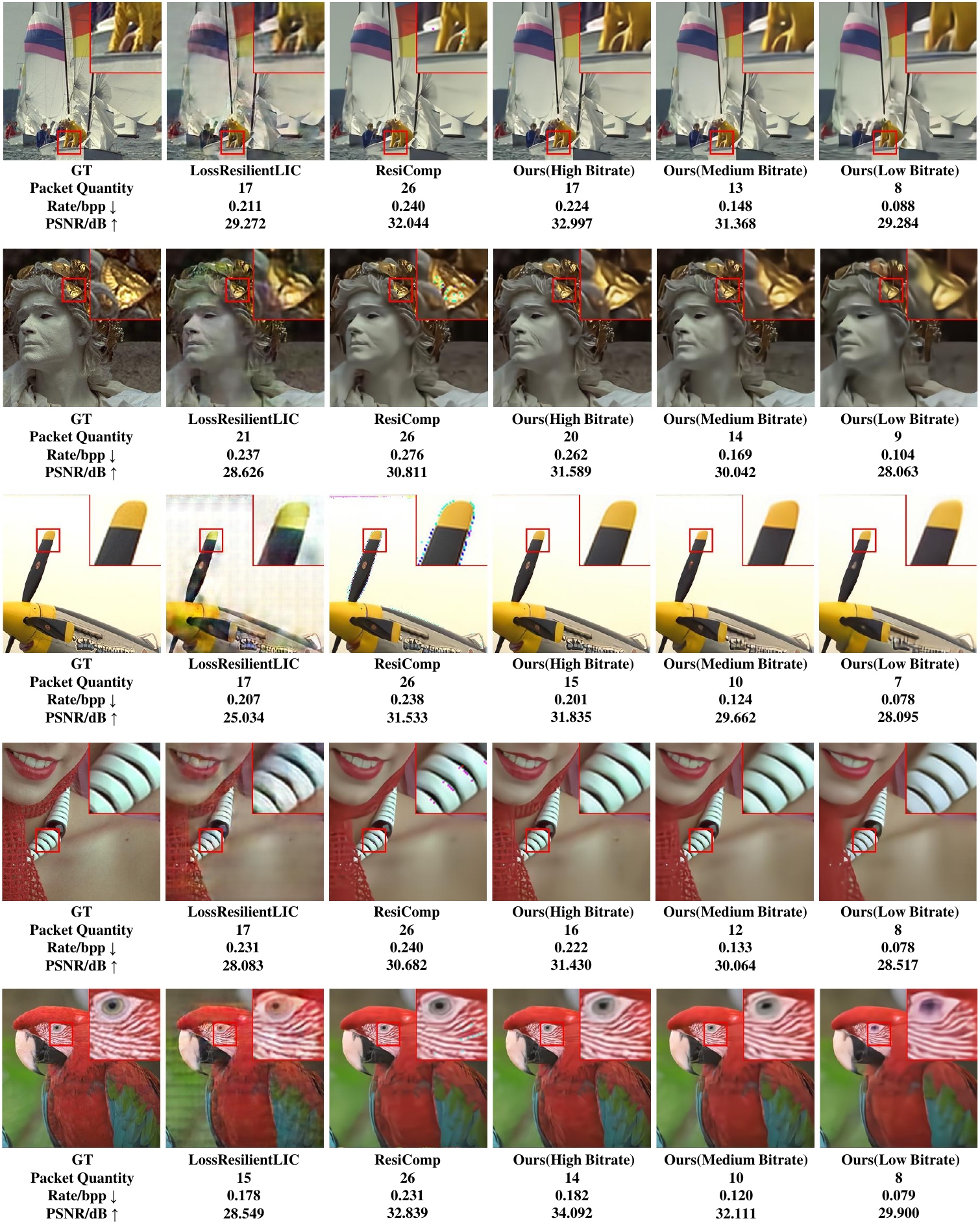}
    \caption{Visualization of reconstructed images on Kodak dataset. All methods lose the first two packets, while in our method, the packet of \(y_3\) is also lost due to the autoregressive dependency. Our method achieves better results at a lower bitrate.}
  \Description{fig:appendix_visionkodak}
  \label{fig:appendix_visionkodak}
\end{figure*}

\begin{figure*}[t]
  \includegraphics[width=\textwidth]{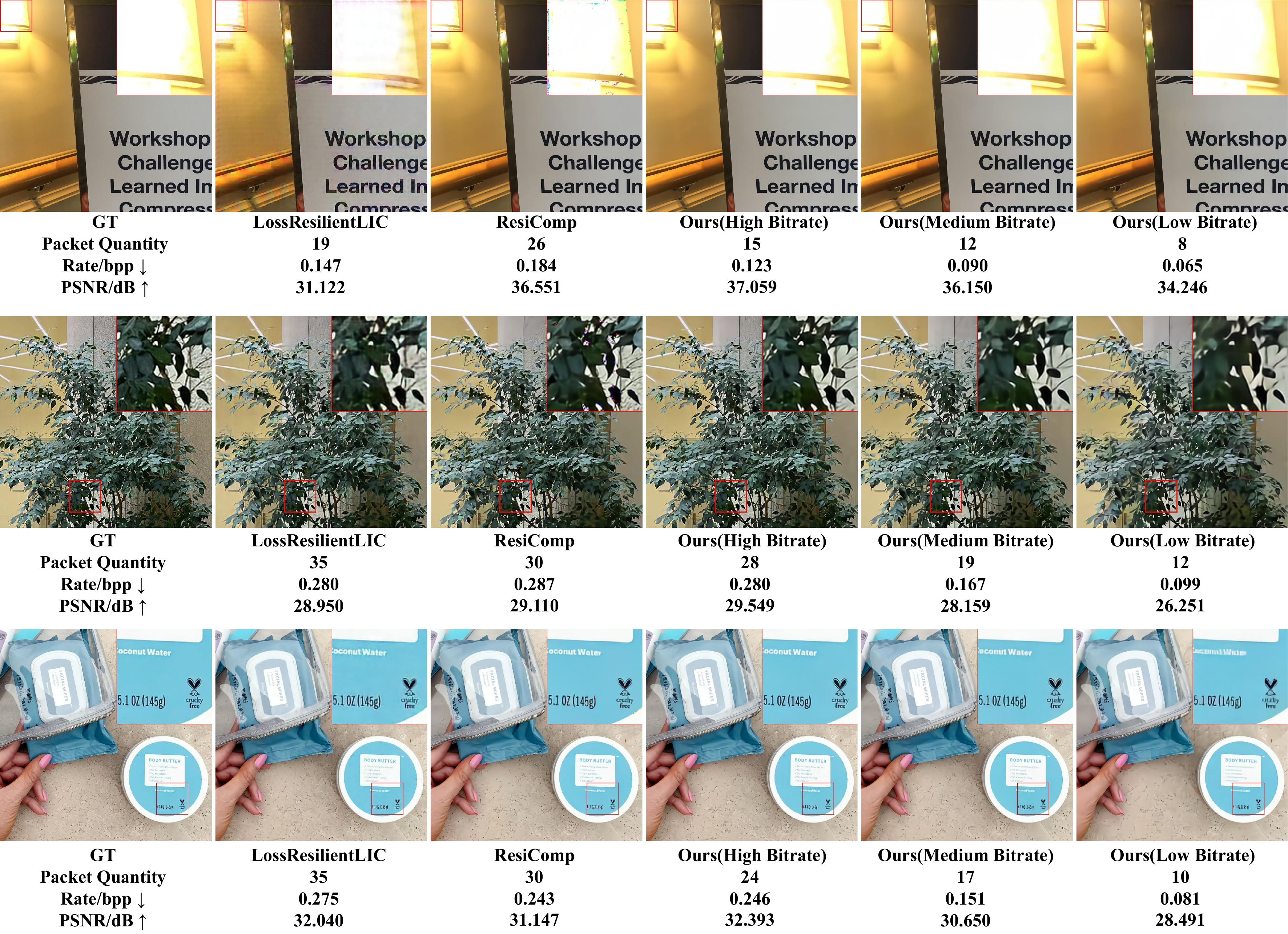}
    \caption{Visualization of reconstructed images on CLIC dataset. All methods lose the first two packets, while in our method, the packet of \(y_3\) is also lost due to the autoregressive dependency. Our method achieves better results at a lower bitrate.}
  \Description{fig:appendix_visionclic}
  \label{fig:appendix_visionclic}
\end{figure*}

More visual comparisons on the Kodak and CLIC datasets are provided in Figure~\ref{fig:appendix_visionkodak} and Figure~\ref{fig:appendix_visionclic}. These additional examples cover diverse image contents and demonstrate the consistency of reconstruction performance across different scenarios. The visual results are consistent with the quantitative evaluation, showing that our method maintains higher perceptual quality under packet loss while effectively suppressing reconstruction artifacts. These results further validate the robustness of the proposed design under challenging transmission conditions.

\begin{table*}[h]
\centering
\caption{Gains over Baseline under packet loss with a uniform distribution on Kodak dataset.}
\label{tab:performance-comparison-kodak-appendix}
\small
\begin{tabular*}{\textwidth}{@{\extracolsep{\fill}} l l *{9}{c}}
\toprule
\multirow{2}{*}{$p_e$} & \multirow{2}{*}{Method} 
& \multicolumn{3}{c}{Low Bitrate} 
& \multicolumn{3}{c}{Medium Bitrate} 
& \multicolumn{3}{c}{High Bitrate} \\
\cmidrule(lr){3-5} \cmidrule(lr){6-8} \cmidrule(lr){9-11}
& 
& bpp & mean $\uparrow$ & $\Delta$ $\uparrow$ 
& bpp & mean $\uparrow$ & $\Delta$ $\uparrow$ 
& bpp & mean $\uparrow$ & $\Delta$ $\uparrow$ \\
\midrule

\multirow{6}{*}{5\%}
& LossResilientLIC Baseline & 0.133 & 26.551 & - & 0.215 & 26.976 & - & 0.336 & 28.619 & - \\
& LossResilientLIC Random   & 0.133 & 26.657 & 0.106 & 0.220& 27.754 & 0.778 & 0.351 & 29.256 & 0.637 \\
& LossResilientLIC          & 0.131 & 26.850  & 0.299 & 0.210& 28.247 & 1.271 & 0.338 & 29.561 & 0.942 \\
& Our Baseline              & 0.1284 & 26.934 & - & 0.216 & 27.013 & - & 0.332 & 29.194 & - \\
& Our Random                & 0.130   & 27.251 & 0.317 & 0.208 & 28.497 & 1.484 & 0.334 & 29.704 & 0.511\\
& Ours                      & 0.125  & \textbf{27.546} & \textbf{0.612} 
                            & 0.217  & \textbf{28.98}  & \textbf{1.967} 
                            & 0.346  & \textbf{30.433} & \textbf{1.239} \\
\addlinespace

\multirow{6}{*}{10\%}
& LossResilientLIC Baseline & 0.133 & 25.131 & - & 0.215 & 25.855 & - & 0.336 & 27.056 & - \\
& LossResilientLIC Random   & 0.133 & 25.858 & 0.727 & 0.221 & 27.320 & 1.465 & 0.351 & 28.569 & 1.513 \\
& LossResilientLIC          & 0.131 & 26.342 & 1.211 & 0.210& 27.605 & 1.750 & 0.342 & 28.801 & 1.745 \\
& Our Baseline              & 0.1284 & 25.946 & - & 0.216 & 26.489 & - & 0.332 & 27.385 & - \\
& Our Random                & 0.130  & 26.819 & 0.873 & 0.208 & 28.103 & 1.614 & 0.334 & 29.131 & 1.746 \\
& Ours                      & 0.125  & \textbf{27.470}  & \textbf{1.524} 
                            & 0.217  & \textbf{28.845} & \textbf{2.356} 
                            & 0.346  & \textbf{30.271} & \textbf{2.886} \\
\addlinespace

\multirow{6}{*}{20\%}
& LossResilientLIC Baseline & 0.133 & 23.257 & - & 0.215 & 22.524 & - & 0.336 & 23.592 & - \\
& LossResilientLIC Random   & 0.133 & 25.217 & 1.960 & 0.228 & 25.886 & 3.362 & 0.358 & 27.620& 4.028 \\
& LossResilientLIC          & 0.131 & 25.578 & 2.321 & 0.210& 26.617 & 4.093 & 0.346 & 27.966 & 4.374 \\
& Our Baseline              & 0.1284 & 23.534 & - & 0.216 & 22.724 & - & 0.332 & 23.753 & - \\
& Our Random                & 0.130  & 26.482 & 2.948 & 0.208 & 27.441 & 4.717 & 0.334 & 28.168 & 4.415 \\
& Ours                      & 0.125  & \textbf{27.211} & \textbf{3.677} 
                            & 0.217  & \textbf{28.561} & \textbf{5.837} 
                            & 0.346  & \textbf{29.931} & \textbf{6.178} \\
\bottomrule
\end{tabular*}
\end{table*}

\begin{table*}[h]
\centering
\caption{Performance comparison of models under packet loss with a uniform distribution on CLIC dataset. Setting 1 corresponds to a packet size of 4500 bytes, and Setting 2 to 900 bytes.}
\label{tab:performance-comparison-clic-appendix}
\small
\begin{tabular*}{\textwidth}{@{\extracolsep{\fill}} l l *{9}{c}}
\toprule
\multirow{2}{*}{$p_e$} & \multirow{2}{*}{Method} 
& \multicolumn{3}{c}{Low Bitrate} 
& \multicolumn{3}{c}{Medium Bitrate} 
& \multicolumn{3}{c}{High Bitrate} \\
\cmidrule(lr){3-5} \cmidrule(lr){6-8} \cmidrule(lr){9-11}
& 
& bpp & mean $\uparrow$ & var $\downarrow$ 
& bpp & mean $\uparrow$ & var $\downarrow$ 
& bpp & mean $\uparrow$ & var $\downarrow$ \\
\midrule
\multirow{8}{*}{5\%} 
& JPEG2000(Setting 1)  & 0.104 & 22.083 & 0.072 & 0.166 & 25.261 & 0.109 & 0.252 & 27.181 & 0.157 \\
& LossResilientLIC(Setting 1) & 0.100 & 28.795  & 0.039 & 0.158 & 30.074 & 0.060 & 0.252 & 31.154 & 0.062 \\
& ResiComp(Setting 1) & 0.124 & \textbf{30.864} & 0.088 & 0.147 & \textbf{32.132} & 0.161 & 0.218 & 32.551 & 0.227 \\
& ResiComp(Setting 2) & 0.126 & 28.104 & 0.142 & 0.147 & 30.552 & 0.297 & 0.269 & 32.550 & 0.116 \\
& Our Baseline  & 0.090 & 29.197 & 0.023 & 0.147 & 30.299 & 0.056 & 0.230 & 31.078 & 0.091 \\
& Our Random(Setting 1)  & 0.091   & 30.106 & 0.003 & 0.148 & 31.096 & 0.003 & 0.225 & 31.736 & 0.005 \\
& Our Random(Setting 2)  & 0.091   & 29.084 & 0.005 & 0.148 & 29.382 & 0.003 & 0.225 & 29.887 & \(5.4 \times 10^{-4}\) \\
& Ours(Setting 1)   & 0.090  & 30.550 & \underline{\(1.1 \times 10^{-4}\)}  & 0.148  & 31.928  & \underline{\(8.9 \times 10^{-5}\)} & 0.228  & \underline{33.262} & \underline{\(3.4 \times 10^{-5}\)} \\
& Ours(Setting 2) & 0.090  & \underline{30.551} & \textbf{\(\textbf{2.8} \times \textbf{10}^\textbf{{-5}}\)}  & 0.148  & \underline{31.931}  & \textbf{\(\textbf{1.6} \times \textbf{10}^{\textbf{-5}}\)} & 0.228  & \textbf{33.282} & \textbf{\(\textbf{1.3} \times \textbf{10}^{\textbf{-5}}\)} \\

\addlinespace
\multirow{8}{*}{10\%} 
& JPEG2000(Setting 1)  & 0.104 & 20.417 & 0.081 & 0.166 & 22.680 & 0.085 & 0.252 & 25.53 & 0.155 \\
& LossResilientLIC(Setting 1) & 0.100 & 28.261  & 0.115 & 0.158 & 29.210 & 0.200 & 0.252 & 30.491 & 0.080 \\
& ResiComp(Setting 1) & 0.124 & 28.693 & 0.112 & 0.147 & \underline{31.531} & 0.177 & 0.228 & 32.263 & 0.167 \\
& ResiComp(Setting 2) & 0.126 & 26.076 & 0.197 & 0.147 & 28.688 & 0.312 & 0.269 & 32.241 & 0.104 \\
& Our Baseline & 0.090 & 27.152 & 0.062 & 0.147 & 27.862 & 0.153 & 0.230 & 28.544 & 0.170 \\
& Our Random(Setting 1) & 0.091   & 29.300 & 0.005 & 0.148 & 30.004 & 0.023 & 0.225 & 30.276 & 0.113 \\
& Our Random(Setting 2) & 0.091 & 27.798 & 0.011 & 0.148 & 28.220 & 0.006 & 0.225 & 29.201 & $2.3\times 10^{-4}$ \\
& Ours(Setting 1) & 0.090  & \underline{30.212} & \underline{\({3.4} \times {10}^{{-4}}\)} & 0.148  & {31.461}  & \underline{\({6.8} \times {10}^{{-4}}\)} & 0.228  & \underline{32.990} & \underline{\({7.0} \times {10}^{{-5}}\)} \\
& Ours(Setting 2) & 0.090  & \textbf{30.382} & \textbf{\(\textbf{3.3} \times \textbf{10}^{\textbf{-5}}\)} & 0.148  & \textbf{31.789}  & \textbf{\(\textbf{5.7} \times \textbf{10}^{\textbf{-5}}\)} & 0.228  & \textbf{33.138} & \textbf{\(\textbf{3.1} \times \textbf{10}^{\textbf{-5}}\)} \\

\addlinespace
\multirow{8}{*}{20\%} 
& JPEG2000(Setting 1)  & 0.104 & 17.033 & 0.092 & 0.166 & 19.698 & 0.278 & 0.252 & 21.87 & 0.203 \\
& LossResilientLIC(Setting 1) & 0.103 & 27.316  & 0.110 & 0.167 & 27.873 & 0.318 & 0.257 & 29.469 & 0.122 \\
& ResiComp(Setting 1) & 0.124 & 26.135 & 0.319 & 0.147 & 30.311 & 0.300 & 0.228 & 31.636 & 0.414 \\
& ResiComp(Setting 2) & 0.126 & 22.164 & 0.273 & 0.147 & 26.447 & 0.417 & 0.269 & 31.611 & 0.254 \\
& Our Baseline & 0.090 & 24.056 & 0.098 & 0.147 & 24.702 & 0.137 & 0.230 & 25.818 & 0.253 \\
& Our Random(Setting 1) & 0.091   & 27.417 & 0.082 & 0.148 & 27.940 & 0.092 & 0.225 & 28.244 & 0.049 \\
& Our Random(Setting 2) & 0.091   & 25.957 & 0.012 & 0.148 & 26.946 & 0.021 & 0.225 & 28.054 & 0.007 \\
& Ours(Setting 1) & 0.090  & \underline{29.862} & \(\underline{2.9 \times {10}^{{-4}}}\) & 0.148  & \underline{31.200}  & \(\underline{2.6 \times {10}^{{-4}}}\) & 0.228  & \underline{32.508} & \(\underline{3.4 \times {10}^{{-4}}}\) \\
& Ours(Setting 2) & 0.090  & \textbf{30.013} & \(\textbf{6.3} \times \textbf{10}^{\textbf{-5}}\) & 0.148  & \textbf{31.449}  & \(\textbf{3.6} \times \textbf{10}^{\textbf{-4}}\) & 0.228  & \textbf{32.785} & \(\textbf{2.2} \times \textbf{10}^{\textbf{-4}}\) \\
\bottomrule
\end{tabular*}
\end{table*}

\end{document}